\documentclass{article} 
\usepackage[final]{colm2026_conference}

\usepackage{microtype}
\usepackage{hyperref}
\usepackage{url}
\usepackage{booktabs}

\usepackage{lineno}

\definecolor{darkblue}{rgb}{0, 0, 0.5}
\hypersetup{colorlinks=true, citecolor=darkblue, linkcolor=darkblue, urlcolor=darkblue}

\usepackage{amsmath}
\usepackage{amssymb}
\usepackage{algorithm}
\usepackage{algorithmic}
\usepackage{xspace}
\usepackage{tabularx}
\usepackage{multirow}
\usepackage{booktabs}
\usepackage{colortbl}
\usepackage{xcolor}
\usepackage[export]{adjustbox}
\usepackage{enumitem}
\usepackage[T1]{fontenc}
\usepackage[breakable,skins,listings]{tcolorbox}

\graphicspath{{./figures/}}

\newcommand{\eg}{\textit{e.g.}\xspace}

\title{Don't Lose the Thread: Empowering Long-Horizon LLM Agents with Cognitive Resource Self-Allocation}

\author{
Zheng Zhu$^{1}$ \quad
Xiaozhuan Liang$^{2}$ \quad
Xi Chen$^{2}$ \quad
Bei Yu$^{1}$ \quad
Jiaya Jia$^{3}$ \\
{\normalfont $^{1}$The Chinese University of Hong Kong
\quad $^{2}$Tencent} \\
{\normalfont $^{3}$The Hong Kong University of Science and Technology}
}

\begin{document}

\ifcolmsubmission
\linenumbers
\fi

\maketitle

\begin{abstract}
Agents powered by large language models (LLMs) have demonstrated remarkable progress in solving complex reasoning tasks. However, LLM agents often falter on long-horizon tasks due to cognitive overload, as their working memory becomes cluttered with expanding and irrelevant information, which dilutes their attention and hinders effective planning and reasoning. To mitigate this challenge, we introduce \textbf{CO}gnitive \textbf{R}esource Self-\textbf{AL}location (\textbf{CORAL}), a novel reasoning paradigm that empowers agents to proactively optimize their context. Implemented as an agent-callable working memory management toolset, CORAL allows an agent to checkpoint its task progress and verified facts, and adaptively initiate a new problem-solving episode by resetting cluttered context while preserving only the checkpointed information, enabling it to resume reasoning from a clean state. We further enhance the agent's checkpointing capabilities using a Multi-episode Agentic Reinforced Policy Optimization algorithm. On the GAIA benchmark, CORAL with an 8B model achieves 42.8\% average accuracy, with particularly strong gains on the most challenging long-horizon tasks (Level~2 and Level~3). Crucially, these gains come at a substantially \emph{lower} inference cost: CORAL consumes 6--7$\times$ fewer input tokens than a ReAct agent. It also generalizes to multi-hop Knowledge QA benchmarks and to the considerably harder BrowseComp, where it improves a strong proprietary agent by 8.0 points.
\end{abstract}

\section{Introduction}

LLM-driven agents extend the capabilities of Large Language Models (LLMs) through the integration of external tools \citep{openaidr, gemini, liu2025scalecua, li2025torl}, substantially outperforming single-turn inference. On long-horizon tasks they operate a \textsc{thought-action-observation} cycle \citep{react2022}, interleaving planning, environmental interaction, and reasoning \citep{erdogan2025plan, qiao2024agent, huang2024understanding}. A critical challenge arises as each cycle populates the context with verbose environmental feedback and a history of failed attempts \citep{wu2025gui, shinn2023reflexion}. This escalating contextual noise progressively degrades the model's planning and reasoning faculties \citep{yang2025llm}, a phenomenon comparable to the cognitive overload that impairs human problem-solving when working memory becomes saturated \citep{baddeley1974working, sweller1988cognitive}.

Current paradigms for context optimization seek to prevent this bloat by pruning messages or distilling salient information, generally activated by rule-based heuristics such as fixed intervals \citep{mem12025, memagent2025} or imminent saturation of the context window \citep{wu2025resum}, and implemented either by truncating the context directly \citep{timrun2025} or by using external models to compress it \citep{wu2025resum}.

\begin{figure*}[ht]
\centering
\includegraphics[width=1.0\linewidth]{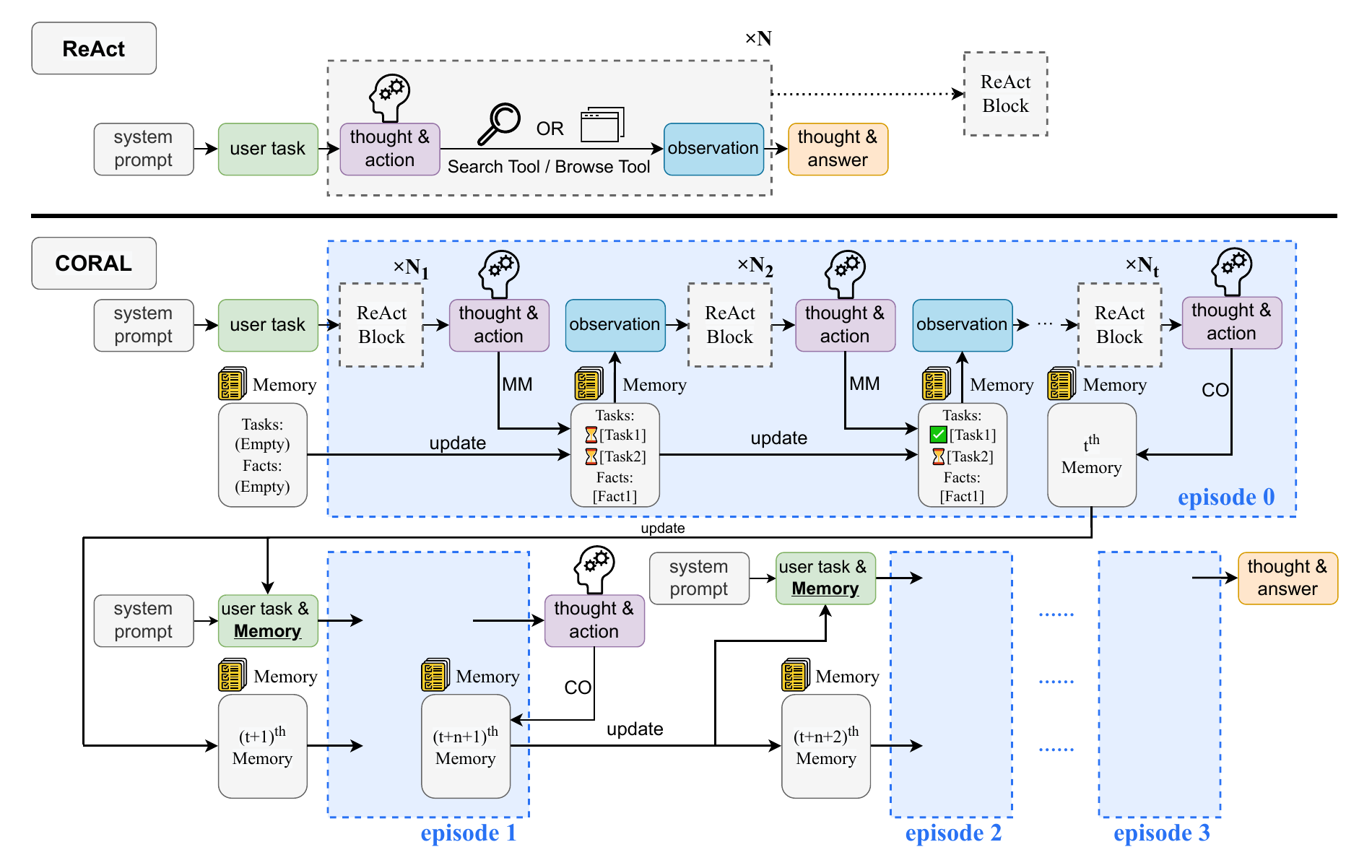}
\caption{Comparison of the ReAct and CORAL frameworks; both rows read left to right as the context grows. \textbf{Top (ReAct).} One context accumulates $N$ \textit{ReAct blocks} (thought, action, observation) above the system prompt and user task, and nothing is ever removed. \textbf{Bottom (CORAL).} The same blocks are interleaved with two agent-callable tools. \textit{Memory Management} (MM) writes into the persistent \textsc{Memory} panel below the trace, holding \textit{task progress} (the plan, each sub-goal pending or completed) and \textit{verified facts}, so that panel evolves while the raw trace only grows. \textit{Context Optimization} (CO), the vertical bar, ends the current \textit{episode}: everything to its left is discarded and the next episode restarts from only the system prompt, the user task, and the current panel. A trajectory is thus a sequence of episodes (here 0--3).}
\label{fig:CORAL}
\end{figure*}

Fundamentally, this redundant context is a consequence of the agent's own imperfect planning: suboptimal tool use generates irrelevant environmental feedback \citep{wang2025otc}, which clutters the context and further degrades reasoning, creating a vicious cycle. One line of research breaks this cycle with agentic reinforcement learning (RL) over the tool-use policy \citep{qian2025toolrl, zhang2025nemotron, jin2025search}, but on long-horizon tasks such methods struggle with reward sparsity, since relying solely on final outcomes makes credit assignment to intermediate actions difficult \citep{gig2025, xia2025agentrm}. Proactive context optimization is a powerful orthogonal approach, improving both reasoning quality and RL training dynamics \citep{kimi_researcher, wu2025resum}.

We therefore introduce the \textbf{CO}gnitive \textbf{R}esource Self-\textbf{AL}location (\textbf{CORAL}) framework, which extends existing agentic architectures with a callable toolset for working memory management. The agent autonomously invokes a \textit{Memory Management} tool to checkpoint its task progress and verified facts, and a \textit{Context Optimization} tool to reset cluttered context while preserving only the checkpointed information, so it can resume reasoning from a clean state. This systematically refocuses the agent's cognitive resources on its current state rather than letting them be diluted by obsolete information. We first instill the checkpointing ability through Supervised Fine-Tuning (SFT).

To further enable the model to discover optimal checkpointing strategies without additional trajectory data, we introduce a Multi-episode Agentic Reinforced Policy Optimization (Multi-episode ARPO) algorithm, which refines the agent's checkpointing policy and improves the overall agentic RL dynamics.

The key contributions of this work are as follows:

\begin{itemize}[leftmargin=1em]
    \item We introduce \textbf{CORAL}, a framework that empowers agents to manage their own working memory through a callable toolset comprising a Memory Management tool and a Context Optimization tool. By dynamically checkpointing progress and resetting cluttered context, CORAL enables robust long-horizon planning and reasoning.
    \item We propose a two-stage training approach: Supervised Fine-Tuning (SFT) to instill core checkpointing skills, followed by \textbf{Multi-episode ARPO}, a novel RL algorithm that propagates trajectory-level rewards across episode boundaries to optimize checkpointing strategies.
    \item On GAIA, a suite of multi-hop Knowledge QA tasks, and the harder BrowseComp benchmark, CORAL significantly outperforms existing approaches on challenging long-horizon tasks---while consuming 6--7$\times$ fewer input tokens than a ReAct agent, so that better attention management is a source of accuracy rather than a cost paid for it.
\end{itemize}

\section{Preliminaries}

\subsection{Problem formulation}
We consider a general large language model (LLM)-based agent that, given a problem specification, interacts with its environment over a sequence of reasoning and action steps.
Following ReAct~\citep{react2022}, these steps are iterations of \textit{Thought}-\textit{Action}-\textit{Observation}. 
Given a system prompt $s$ and a question $q\in p(Q)$, the agent $\pi_\theta$ at step $t$ generates \textit{Thought} $r_t \sim \pi_\theta(\cdot|c_t)$ and \textit{Action} $a_t \sim \pi_\theta(\cdot|c_t,r_t)$, where $c_t=(s, q, r_1, a_1, o_1, \dots, r_{t-1}, a_{t-1}, o_{t-1})$ is the context at step $t$; the environment then returns \textit{Observation} $o_t$.
The loop ends when the agent solves the question or reaches the max steps, so an episode of $M$ steps is $e_\text{terminated} = (s, q, r_1, a_1, o_1, \dots, r_M, a_M, o_M)$ if it exhausts its budget, and
\begin{equation}
    e_\text{completed} = (s, q, r_1, a_1, o_1, \dots, r_M)
\end{equation}
if it finishes, where the final \textit{Thought} $r_M$ contains the answer and the episode stops immediately.


\section{\textbf{CO}gnitive \textbf{R}esource Self-\textbf{AL}location (\textbf{CORAL})}

Our design is inspired by cognitive psychology's treatment of problem solving as the management of finite resources. \citet{kahneman1973attention} models attention as a limited-capacity resource allocated across competing demands and \citet{baddeley1974working} characterize working memory as a small, actively maintained store supporting ongoing reasoning; cognitive load theory \citep{sweller1988cognitive} explains why overloading that store is harmful, as performance degrades once the material a solver must hold exceeds capacity, even when no individual step is difficult. The standard human remedy is \textit{cognitive offloading} \citep{risko2016cognitive}: externalizing intermediate state, for instance by writing it down, so internal capacity is spent on the current sub-problem rather than on retention.

The parallel to LLM agents is direct: the agent's context serves as its working memory, and on long-horizon tasks it becomes progressively cluttered with intermediate steps, leading to cognitive overload. We therefore introduce \textbf{CO}gnitive \textbf{R}esource Self-\textbf{AL}location (\textbf{CORAL}), the agentic analogue of cognitive offloading, in which the agent consolidates progress and refocuses attention by writing checkpoints into an external store and purging the context those checkpoints summarize.

\subsection{Working memory management toolset}
We operationalize this through two agent-callable tools. Critically, both are invoked autonomously by the agent from its own judgment of the task state, rather than triggered by rule-based heuristics or external controllers.
\begin{itemize}[leftmargin=1em]
    \item \textbf{Memory Management (MM).} The agent's checkpointing mechanism, maintaining two categories of persistent information in working memory. \textit{Task progress} captures the high-level plan and tracks which sub-goals are complete, enabling coherent multi-step planning; \textit{verified facts} stores key information the agent has confirmed while reasoning, such as intermediate results or discovered entities. The agent can add, modify, or delete entries in either category, creating checkpoints that survive context resets. In practice it invokes MM at three characteristic moments: at the start of a task to record a plan, on discovering critical information to store it as a verified fact, and after completing a sub-task to update its progress.
    \item \textbf{Context Optimization (CO).} When the agent judges its context excessively long or cluttered with obsolete information, it invokes this tool for a hard reset: the conversational history is cleared and the current working memory is returned as an observation. Formally, when $a_t=a_{CO}$ the tool returns the memory state as $o_t$ and the next episode's initial context is $(s, q, o_t)$, so only checkpointed information survives while accumulated tool outputs and reasoning traces are discarded.
\end{itemize}

\noindent \textbf{Multi-episode trajectory.}
The CO reset segments what would otherwise be a single continuous reasoning process into multiple shorter episodes, each starting from a refreshed context that retains only essential information.
To capture this behavior, we extend the single-episode formulation to a \textit{multi-episode trajectory}.
We assume that there are $N$ episodes in total, then the i-th episode (with $M_i$ iterations) can be formulated as:
\begin{align}
    \label{eq:episode}
    & e_i = \left\{
        \begin{array}{ll}
            \begin{aligned}
                (c_i, r_{i,1}, a_{i,1}, o_{i,1}, \dots, \\
                 r_{i,M_i}, a_{i,M_i}, o_{i,M_i}),
            \end{aligned} & i<N \\
            (c_i, r_{i,1}, a_{i,1}, o_{i,1}, \dots, r_{i,M_i}), & i=N
        \end{array}
    \right. && \\
    \label{eq:context}
    & \text{where} \quad c_i = \left\{
        \begin{array}{ll}
            (s, q), & i=1 \\
            (s, q, o_{{i-1},M_{i-1}}), & i>1
        \end{array}
    \right. &&
\end{align}
In the final episode the last thought $r_{N,M_N}$ contains the answer and the episode ends immediately. The initial context $c_i$ is $(s,q)$ for the first episode, and for subsequent ones adds the memory state $o_{{i-1},M_{i-1}}$ returned by the \textit{Context Optimization} tool of the previous episode.
A complete trajectory with $N$ episodes is then:
\begin{align}
    \label{eq:traj}
    \mathcal{T} = (e_1, e_2, ..., e_N)
\end{align}

\subsection{CORAL framework}
As illustrated in~\autoref{fig:CORAL}, CORAL augments the traditional ReAct framework with the toolset above. Within an episode the agent interleaves reasoning steps with MM invocations to incrementally build its checkpoint; when it judges the context sufficiently cluttered it invokes CO to start a fresh episode carrying only the distilled checkpoint. This checkpoint-and-reset cycle sustains effective reasoning across arbitrarily long horizons.


\subsection{Further enhancement methods}
While the CORAL framework can be implemented in a prompting-only fashion, we explore dedicated training methods to further enhance its capabilities.

\noindent \textbf{Behavior Cloning.}
To endow the agent with basic function call ability, we apply behavior cloning through supervised fine-tuning (SFT) on curated, high-quality trajectories. 
From~\autoref{eq:traj} we know that a trajectory consists of multiple context-independent episodes, therefore, we split the trajectory into episodes, and fine-tune the model using batches of episodes.
We stress that this \textit{splits} one coherent trajectory rather than \textit{stitching} unrelated ones together: every episode $e_i$ in $\mathcal{T}$ comes from a single model solving a single question in one uninterrupted run, and $e_{i+1}$ begins from the memory state that $e_i$ itself wrote (\autoref{eq:context}). Splitting is necessary precisely because episodes have different initial contexts---the first starts from $(s,q)$, later ones from $(s,q,o_{i-1,M_{i-1}})$---so presenting them as one flat conversation would misrepresent what the agent sees at inference time. Appendix~\ref{sec:data-walkthrough} walks through a complete trajectory and its training instances.
For each episode described in \autoref{eq:episode} and \autoref{eq:context}, we compute the loss using the following loss function:
\begin{equation}
    \label{eq:sft}
    L =  -\frac{1}{\sum_{i=1}^{|e|} \mathbb{I}(x_i \ne o)} \sum_{i=1}^{|e|} \mathbb{I}(x_i \ne o) \cdot \log \pi_{\theta}(x_i \mid x_{<i})
\end{equation}
where $\mathbb{I}(\cdot)$ is the indicator function. Here $\mathbb{I}(x_i \ne o)$ masks out the loss from observation tokens, ensuring the loss is computed over the agent's own generated outputs, such as its reasoning steps (thoughts) and function calls (actions).
By doing so, we only supervise the model on the behaviors it is expected to learn, rather than penalizing it for failing to predict external information from the environment.

An episode's position within its trajectory determines which skill it supervises, and this decomposition is what the multi-episode format buys us. Non-terminal episodes teach the model to progress \textit{while} maintaining its working memory and to recognize when a batch of sub-tasks is complete and the context should be discarded, so the CO call ending such an episode is a supervised decision about \textit{when} to reset rather than a rule fired by an external controller. Later episodes teach resuming from a checkpoint: the model is handed a compact progress-and-facts state instead of a history and must locate the next sub-task without redoing completed work. A ReAct agent trained on flat trajectories encounters neither situation, which is why these skills cannot come from conventional single-episode agentic SFT.

\begin{figure*}[t]
\centering
\includegraphics[width=\linewidth]{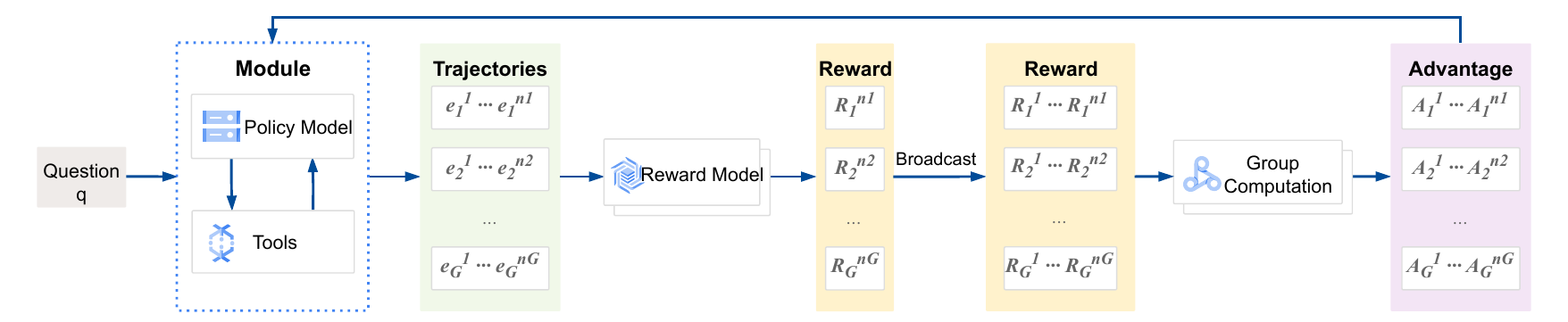}
\caption{Multi-episode ARPO. The reward is computed from a trajectory's final episode (which contains the answer) and propagated to all preceding ones; episodes from all sampled trajectories are then pooled for group-based advantage estimation, enabling credit assignment across episode boundaries.}
\label{fig:me-dapo}
\end{figure*}

\noindent \textbf{Multi-episode Agentic Reinforced Policy Optimization.}
Standard agentic RL methods like DAPO~\citep{yu2025dapo} optimize single-episode trajectories, whereas in CORAL a trajectory consists of multiple episodes separated by context optimization (\autoref{eq:traj}). We extend DAPO to multi-episode trajectories by treating each episode as a separate optimization unit while propagating trajectory-level rewards.

~\autoref{fig:me-dapo} illustrates the idea. Consider sampling $G$ multi-episode trajectories $\{\mathcal{T}_i\}_{i=1}^G$ for a question $q$, where $\mathcal{T}_i = (e_i^1, e_i^2, ..., e_i^{N_i})$ contains $N_i$ episodes. Since all episodes within a trajectory contribute to the final outcome, they should share the trajectory-level reward: we compute $R_i$ from the final episode $e_i^{N_i}$ (which contains the answer) and propagate it backwards,
\begin{equation}
    R_i^j = R_i, \quad \forall j \in \{1, 2, ..., N_i\}
    \label{eq:reward_propagation}
\end{equation}
then normalize over all episodes for the same question and pool them for batch optimization. Letting $\mathcal{E} = \{e_i^j \mid i \in [1,G], j \in [1, N_i]\}$ denote the set of all episodes, the Multi-episode ARPO objective is:

\begin{small}
\begin{equation}
\begin{aligned}
\mathcal{J}_{\text{ME-ARPO}}(\theta) = \mathbb{E}_{q\sim \mathcal{D}, \{\mathcal{T}_i\}_{i=1}^G\sim \pi_{\theta_\text{old}}}
\Bigg[&\frac{1}{\sum_{e \in \mathcal{E}}|e|}\sum_{e_i^j \in \mathcal{E}}\sum_{t=1}^{|e_i^j|} \\
&\min \Big( r_{i,j,t}(\theta) \hat{A}_{i}^{j},  
\ \text{clip} \Big( r_{i,j,t}(\theta), 1 - {\varepsilon_{\text{low}}}, 1 + {\varepsilon_{\text{high}}} \Big) \hat{A}_{i}^{j} \Big) \Bigg]
\label{eq:mearpoloss}
\end{aligned}
\end{equation}
\end{small}
where the probability ratio and advantage are computed as:
\begin{small}
\begin{equation}
    r_{i,j,t}(\theta)=\frac{\pi_{\theta}(e_{i,t}^j \mid c_i^j, e_{i,<t}^j)}{\pi_{\theta_{\text{old}}}(e_{i,t}^j \mid c_i^j, e_{i,<t}^j)},\quad\hat{A}_{i}^{j} = \frac{R_i - \text{mean}(\mathcal{R})}{\text{std}(\mathcal{R})}
\label{eq:me_advantage}
\end{equation}
\end{small}
Here $c_i^j$ is the initial context of episode $e_i^j$ (\autoref{eq:context}) and $e_{i,t}^j$ its $t$-th token. The multiset $\mathcal{R} = \{R_{i,j} \mid i \in [1,G], j \in [1, N_i]\}$ contains the rewards of all episodes for question $q$, each inheriting its parent trajectory's reward, and $\hat{A}_{i}^{j}$ is normalized per question over all episodes. All episodes within a trajectory therefore share one advantage signal, which is what enables credit assignment across episode boundaries and lets the model learn effective checkpointing strategies.



\noindent \textbf{Reward Design.}
We design a simple reward function that consists of a format reward $R_i^{format}$ and an answer reward $R_i^{answer}$. The format reward verifies whether the whole trajectory follows the predefined format and all tool calls in JSON format are valid. The answer reward uses an LLM as a judge to determine whether the final answer is correct.
\begin{equation}
    R_i=R_i^{format}\times R_i^{answer}
\end{equation}

\section{Experiments}

\subsection{Experimental Settings}
\label{sec:exp-settings}

\noindent \textbf{Agent Setup.} Our agent operates in a ReAct-style loop with two environment tools---\textit{Web Search} and \textit{Web Browse}---alongside the two CORAL tools (MM and CO); Appendix~\ref{sec:tool-definitions} gives full specifications.

\noindent \textbf{Baselines.} We compare against three paradigms: (1) \textit{Direct Inference}: GPT-4o~\citep{gpt4o}, DeepSeek-R1~\citep{r1}, Claude Sonnet 4~\citep{claude4}; (2) \textit{Agentic Inference}: R1-Searcher~\citep{r1searcher2025}, WebDancer~\citep{webdancer2025}, WebThinker~\citep{webthinker2025}, Cognitive Kernel-Pro~\citep{fang2025cognitive}, WebSailor~\citep{li2025websailor}, WebShaper~\citep{tao2025webshaper}, OpenAI Deep Research~\citep{openaidr}; (3) \textit{ReAct}: the classic paradigm using web search and browse tools. The agentic baselines are themselves trained systems of comparable complexity---multi-stage RL, SFT with iterative DPO, or SFT with RL---so \autoref{tab:gaia-results} compares trained agents with different architectures, not a trained CORAL against untrained alternatives.

\noindent \textbf{CORAL Variants.} We evaluate three configurations: \textit{Prompting}, the framework through prompting alone; \textit{SFT}, adding Supervised Fine-Tuning on curated trajectories; and \textit{RL}, adding Multi-episode ARPO on top of SFT.

\noindent \textbf{Benchmarks.} We evaluate on (1) \textbf{GAIA}~\citep{gaia2023}, using the standard 103-sample text-only validation subset (39 Level~1, 52 Level~2, 12 Level~3);
(2) \textbf{Knowledge QA}, a suite of multi-hop benchmarks: Natural Questions~\citep{kwiatkowski2019natural}, HotpotQA~\citep{yang2018hotpotqa}, 2WikiMultiHopQA~\citep{ho2020constructing}, Musique~\citep{trivedi2022musique}, and Bamboogle~\citep{press2023measuring};
and (3) \textbf{BrowseComp}~\citep{bc_en2025}, whose questions are built to be unanswerable without many hops of web exploration and which therefore stresses context management hardest.

\noindent \textbf{Dataset.} We follow WebShaper~\citep{tao2025webshaper} to construct questions of controllable difficulty. WebShaper is \textit{formalization-driven}: instead of collecting pages and then asking an LLM to write questions about them, it represents each task as \textit{Knowledge Projections}---set-theoretic constraints over entities and relations---and layer-wise expands a seed question's leaf constants into sub-questions, validating each expansion against retrieved evidence. Reasoning depth thus becomes explicit and controllable (Appendix~\ref{sec:webshaper}).

Claude Sonnet 4, equipped with the CORAL toolset, synthesizes 2,932 web agent trajectories for these questions, of which 1,642 (56\%) reach the correct answer under the LLM-as-a-Judge protocol of Appendix~\ref{sec:llm-judge-prompt} and 1,290 (44\%) do not. We retain both. The incorrect ones are not curated ``near-misses'' but ordinary failures of a capable model---valid reasoning derailed by ambiguous content, a late inference error, or incomplete retrieval---filtered only for well-formedness. We add 241 mathematical reasoning trajectories from NuminaMath~\citep{li2024numinamath} and OpenMathReasoning~\citep{moshkov2025aimo2} to strengthen computation beyond retrieval. The combined 3,173 trajectories expand into 3,682 SFT conversations after episode splitting (Appendix~\ref{sec:data-walkthrough}); the RL stage uses web tasks only, and all training questions are synthesized independently of every test set. Hyperparameters are in Appendix~\ref{sec:training-details}.

\noindent \textbf{Evaluation Protocol.} We sample 8 independent trajectories per question with temperature 0.7 and top-p 0.8. Unless otherwise specified we report Pass@1 (Avg@8), the average accuracy across the 8 runs. Pass@8, Maj@8, and standard deviations are in Appendix~\ref{sec:eval-metrics}.

%

\begin{table*}[t]
    \centering
    \small
    \renewcommand{\arraystretch}{1.0} 
    \begin{tabularx}{\linewidth}{
    >{\arraybackslash}p{2.9cm}
    >{\centering\arraybackslash}X
    >{\centering\arraybackslash}X
    >{\centering\arraybackslash}X
    >{\centering\arraybackslash}X
    }
    \toprule
    \textbf{Model}
    & Level 1 & Level 2 & Level 3 & Average \\
    \midrule
    \multicolumn{5}{c}{\cellcolor{gray!20}\textsc{Direct Inference}} \\
    \midrule
    GPT-4o & 23.1 & 15.4 & 8.3 & 17.5 \\
    DeepSeek-R1 & 43.6 & 26.9 & 8.3 & 31.1 \\
    Claude Sonnet 4 & 38.5 & 36.5 & 8.3 & 34.0 \\
    \midrule
    \multicolumn{5}{c}{\cellcolor{gray!20}\textsc{Agentic Inference}} \\
    \midrule
    \multicolumn{5}{l}{\textit{7--8B backbones}} \\
    R1-Searcher-7B & 28.2 & 19.2 & 8.3 & 20.4 \\
    WebDancer-7B & 41.0 & 30.7 & 0.0 & 31.0 \\
    WebSailor-7B & - & - & - & 37.9 \\
    CK-Pro-8B & \textbf{56.4} & 42.3 & 8.3 & \textbf{43.7} \\
    \midrule
    \multicolumn{5}{l}{\textit{$\geq$32B or proprietary backbones}} \\
    WebDancer-32B & 46.1 & 44.2 & 8.3 & 40.7 \\
    WebThinker-32B-RL & 56.4 & 50.0 & 16.7 & 48.5 \\
    WebSailor-72B & - & - & - & 55.4 \\
    WebShaper-72B & - & - & - & 60.1 \\
    OpenAI DR & \underline{74.3} & \textbf{69.1} & \textbf{47.6} & \textbf{67.4} \\
    \midrule
    \multicolumn{5}{c}{\cellcolor{blue!15} \textsc{CORAL (Ours)}} \\
    \midrule
    ReAct \includegraphics[scale=0.03,valign=c]{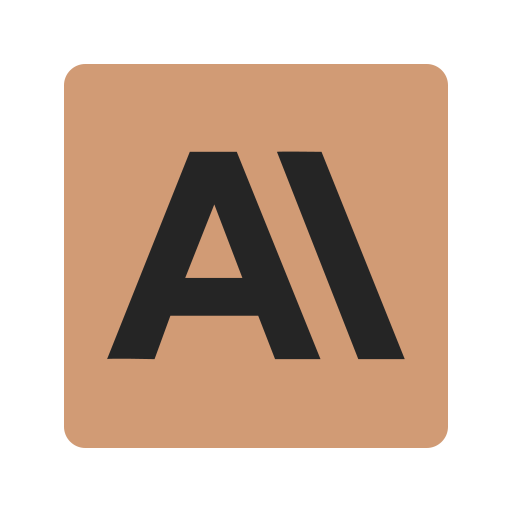}  & 69.2 & 61.5 & 25.0 & 60.2  \\
    \hspace{1em}+\textsc{Prompting} & \textbf{74.4} & \underline{65.4} & \underline{33.3} & \underline{65.0} \\
    \midrule
    ReAct \includegraphics[scale=0.03, valign=c]{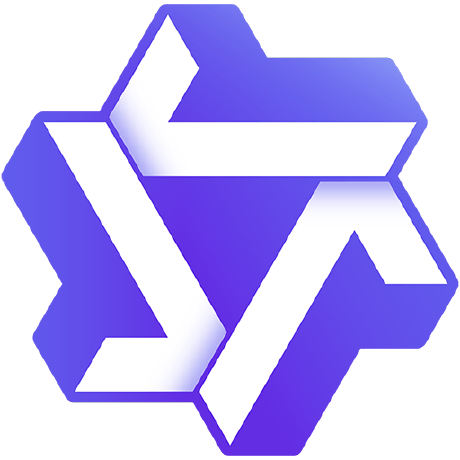} & 33.3 & 11.5 & 8.3 & 19.4 \\
    \hspace{1em}+\textsc{Prompting} & 28.2 & 19.2 & 0.0 & 20.4 \\
    \hspace{1em}+\textsc{SFT} & \underline{47.1} & \underline{43.5} & \underline{15.6} & 41.6 \\
    \hspace{1em}+\textsc{RL} & 46.2 & \textbf{45.7} & \textbf{19.8} & \underline{42.8} \\
    \bottomrule
    \end{tabularx}
    \caption{Main results on GAIA (Pass@1, Avg@8). \textbf{Marking convention:} rows are compared within \emph{parameter-size brackets} and across all three paradigms, not within paradigm blocks. The 7--8B bracket holds the four 7--8B agentic systems and the Qwen3-8B CORAL rows; the other holds everything $\geq$32B or of undisclosed proprietary size, including all \textsc{Direct Inference} rows and the Claude Sonnet 4 CORAL rows. Per bracket and column we \textbf{boldface} the best and \underline{underline} the second best, so each column carries two bold entries and the best need not lie in the CORAL block.}
    \label{tab:gaia-results}
\end{table*}

\subsection{Overall Performance}
\label{sec:overall-performance}

\noindent ~\autoref{tab:gaia-results} presents our main results. CORAL improves substantially over existing methods, particularly on the challenging long-horizon tasks (Level 2 and Level 3) that require extended reasoning chains. Applied to a powerful proprietary model (Claude Sonnet 4), prompting-only CORAL reaches an average of 65.0, comparable with OpenAI Deep Research (67.4), so the approach is effective even on already capable models.

\noindent \textbf{Where the gains lie.} Among 8B-scale models, CK-Pro-8B attains the highest overall score (43.7\% vs.\ our 42.8\%), but the decomposition is instructive: its advantage sits entirely on Level~1 (56.4\% vs.\ 46.2\%), while CORAL with RL leads on Level~2 (45.7\% vs.\ 42.3\%) and far exceeds it on Level~3 (19.8\% vs.\ 8.3\%). Maintaining a working memory costs overhead on questions short enough that clutter never became binding, so a single average understates the method on exactly the regime it targets.

\noindent \textbf{Impact of Training.} On Qwen3-8B the prompting-only CORAL improves only marginally, because of a mismatch between the sophistication of CORAL's tools and the limited agentic capability of the base model, which frequently violates the required format or errs during tool calling. Behavior cloning mitigates this, lifting the model from 20.4\% to 41.6\% on only 3,173 trajectories, and Multi-episode ARPO then reaches 42.8\%. That 1.2-point headline gap understates the RL stage twice over: the gain concentrates where it matters ($+$2.2 on Level~2, $+$4.2 on Level~3, against a slight Level~1 regression), and RL stabilizes the agent, cutting run-to-run standard deviation from 3.5\% to 1.9\% and raising the worst of eight runs from 36.9\% to 39.8\% (Appendix~\ref{sec:eval-metrics}).

\subsection{Inference Efficiency and Tool Usage}
\label{sec:efficiency}


\begin{table}[t]
    \centering
    \small
    \renewcommand{\arraystretch}{1.0}
    \begin{tabularx}{\linewidth}{
    >{\arraybackslash}p{2.75cm}
    >{\centering\arraybackslash}X
    >{\centering\arraybackslash}X
    >{\centering\arraybackslash}X
    >{\centering\arraybackslash}X
    >{\centering\arraybackslash}X
    >{\centering\arraybackslash}X
    }
    \toprule
    \textbf{Method} & \multicolumn{5}{c}{\textit{per trajectory}} & \multirow{2}{*}{GAIA} \\
    \cmidrule(lr){2-6}
    & In.\ tok. & Calls & Peak ctx. & MM & CO & \\
    \midrule
    ReAct           & 1{,}078K & 45.3 & 30.6K & --   & --   & 19.4 \\
    CORAL Prompting & 314K & 21.8 & 14.3K & 20.6 & 4.3  & 20.4 \\
    CORAL SFT       & 153K & 15.9 & 13.9K & 3.9  & 0.21 & 41.6 \\
    CORAL RL        & \textbf{136K} & \textbf{15.2} & \textbf{12.7K} & 4.1 & 0.21 & \textbf{42.8} \\
    \bottomrule
    \end{tabularx}
    \caption{Inference cost and working-memory tool usage on GAIA, over all 824 evaluation trajectories (103 questions $\times$ 8 samples) with Qwen3-8B. Columns report per-trajectory averages of input tokens, LLM calls, the longest context the model is ever conditioned on (``peak ctx.''), and MM and CO invocations. CORAL RL uses 87.4\% fewer input tokens than ReAct while being more than twice as accurate.}
    \label{tab:efficiency}
\end{table}

Improving accuracy by adding tools invites the question of what it costs, so we instrumented every GAIA evaluation trajectory. \autoref{tab:efficiency} shows CORAL is not more expensive than the ReAct agent it replaces but dramatically cheaper: 136K input tokens per trajectory against 1,078K, over 15.2 LLM calls rather than 45.3. Two mechanisms compound. CO resets bound how long the context may ever become, holding the peak at 12.7K tokens against 30.6K; since a ReAct loop re-reads its whole history every turn, its cost grows quadratically in turns and capping the context attacks the dominant term. Less directly but more powerfully, an agent that has written down its plan and verified facts stops re-searching what it already knows, cutting turns threefold.

The tool-usage columns raise a follow-up: if CO is the mechanism, why does the trained agent invoke it only 0.21 times per trajectory, against the prompting-only agent's 4.3? Training teaches the model not to reset more but to checkpoint better, and an agent whose MM calls capture the right state never accumulates the clutter that would justify a reset---which is just as well, since a reset also discards the history that lets an agent repair an earlier mistake (Appendix~\ref{sec:memory-robustness}).

\subsection{Generalization: Knowledge QA and BrowseComp}
\label{sec:additional-eval}

To evaluate generalizability beyond web-based reasoning, we experiment on a suite of multi-hop Knowledge QA benchmarks, using \textit{the same} Qwen3-8B RL checkpoint evaluated on GAIA---trained only on synthesized web questions plus math, with no exposure to these datasets---and evaluate all methods under the same open web access setting. As~\autoref{tab:knowledge-qa-results} shows, CORAL performs strongly throughout, particularly on Musique (43.2\%), improving consistently over contemporaneous works such as Search-o1~\citep{li2025search}, R1-Searcher~\citep{r1searcher2025}, DeepResearcher~\citep{zheng2025deepresearcher}, and SimpleDeepSearcher~\citep{sun2025simpledeepsearcher}. The checkpointing skill acquired during training is thus transferable rather than tied to the training distribution.


\begin{table}[t]
    \centering
    \small
    \renewcommand{\arraystretch}{1.0} 
    \begin{tabularx}{\linewidth}{
    >{\arraybackslash}p{2.95cm}
    >{\centering\arraybackslash}X
    >{\centering\arraybackslash}X
    >{\centering\arraybackslash}X
    >{\centering\arraybackslash}X
    >{\centering\arraybackslash}X
    >{\centering\arraybackslash}X
    }
    \toprule
    \textbf{Method}
    & NQ & HotpotQA & 2Wiki. & Musique & Bamboogle \\
    \midrule
    Search-o1 & 55.1 & 42.4 & 37.7 & 19.7 & 53.6 \\
    R1-Searcher & 52.3 & 53.1 & 65.8 & 25.6 & 65.6 \\
    DeepResearcher & \underline{61.9} & \underline{64.3} & 66.6 & 29.3 & 72.8 \\
    SimpleDeepSearcher & - & - & \textbf{79.8} & \underline{29.4} & \underline{76.8} \\
    \midrule
    \rowcolor{blue!15} CORAL & \textbf{71.0} & \textbf{70.4} & \underline{79.2} & \textbf{43.2} & \textbf{79.2} \\
    \bottomrule
    \end{tabularx}
    \caption{Results on Knowledge QA benchmarks. We report accuracy (\%) on multi-hop QA tasks.}
    \label{tab:knowledge-qa-results}
\end{table}

This suite establishes transfer but weakly tests our specific claim, since its benchmarks reward retrieval and aggregation more than they punish context clutter. We therefore also evaluate on \textbf{BrowseComp}~\citep{bc_en2025}, whose questions cannot be answered without many hops of exploration. Full evaluation being computationally prohibitive, we report the first 150 questions with Claude Sonnet 4, comparing CORAL against a ReAct agent with identical search and browse tools so that the only difference is the availability of MM and CO. CORAL reaches \textbf{30.7\%} against ReAct's \textbf{22.7\%}: an 8.0-point gain, the largest of any benchmark we measure, with no training whatsoever. Appendix~\ref{sec:browsecomp-case} traces one such question, on which ReAct commits to a wrong entity early and can no longer locate that faulty premise beneath forty-two search results, while CORAL---having checkpointed each entity as it was verified---recovers.

Prompting-only CORAL across five backbones from 7B to proprietary scale (Appendix~\ref{sec:backbone}) shows the benefit tracks capability rather than size: the strongest gain most (QwQ-32B $+$5.8, Claude Sonnet 4 $+$4.8), Qwen3-8B is neutral, and Qwen2.5-7B-Instruct \textit{loses} 3.9 points because it cannot reliably emit JSON for four tools instead of two---an obstacle training removes, with SFT on Qwen2.5-7B reaching 34.4\% against its 17.5\% ReAct baseline.

\subsection{Ablation Study}
\label{sec:ablation}


\begin{table}[t]
    \centering
    \small
    \renewcommand{\arraystretch}{1.0}
    \begin{tabularx}{\linewidth}{
    >{\arraybackslash}p{4.3cm}
    >{\centering\arraybackslash}p{1.3cm}
    >{\centering\arraybackslash}X
    >{\centering\arraybackslash}X
    >{\centering\arraybackslash}X
    >{\centering\arraybackslash}X
    }
    \toprule
    \textbf{SFT training set} & \textbf{\#Conv.} & L1 & L2 & L3 & Avg \\
    \midrule
    Full (web + math)                & 3{,}682 & 47.1 & \textbf{43.5} & 15.6 & \textbf{41.6} \\
    \quad w/o math                   & 3{,}441 & 47.1 & 40.4 & 14.6 & 39.9 \\
    \quad w/o incorrect              & 2{,}030 & \textbf{49.7} & 37.5 & 13.5 & 39.3 \\
    \quad random subsample           & 2{,}030 & 42.0 & 36.8 & \textbf{17.7} & 36.5 \\
    \bottomrule
    \end{tabularx}
    \caption{Ablations on GAIA (Pass@1, Avg@8), all with Qwen3-8B and SFT only. \textbf{\#Conv.} counts SFT conversations \emph{after} episode splitting; 3,173 raw trajectories yield 3,682 because each CO invocation contributes one extra. Each variant is a \emph{filtered} subset of one pool, not a fresh sampling run; ``random subsample'' is a volume-matched control drawn uniformly from the full pool, which isolates data quality from data volume.}
    \label{tab:ablation}
\end{table}

\autoref{tab:ablation} isolates the contribution of each part of our training data. All rows use SFT only, so differences are attributable to data rather than RL.

\noindent \textbf{Mathematical reasoning data.}
Removing the 241 math trajectories costs 1.7 points (41.6\% $\rightarrow$ 39.9\%), concentrated on Level~2 (43.5\% $\rightarrow$ 40.4\%), where questions need multi-step deduction rather than lookup. The math data is entirely out of domain---no browsing, no tool calls, no memory management---so what transfers is the reasoning itself: realistic queries often require intermediate computation, which the web trajectories do not teach.

\noindent \textbf{Incorrect trajectories, at matched volume.}
Comparing ``full'' against ``w/o incorrect'' confounds data quality with volume, since discarding failures also shrinks the set from 3,682 to 2,030 conversations. A volume-matched control---a uniform random subsample at 2,030---separates the two cleanly. At equal volume correct-only data is better (39.3\% vs.\ 36.5\%), so correct trajectories genuinely are higher-quality instances one at a time; yet the full pool beats correct-only by a further 2.3 points despite the added 1,652 conversations being predominantly failures. The reconciliation is that behavior cloning on masked agent outputs supervises the process rather than the conclusion: a trajectory that plans sensibly and checkpoints diligently before misidentifying an entity on its last hop still demonstrates every skill we are installing.

\section{Related Work}

\noindent \textbf{Reinforcement Learning for LLM Agents.}
Unlike supervised fine-tuning, which depends on pre-collected expert data, reinforcement learning (RL) lets agents improve through interaction, and is therefore central to acting in dynamic, open-ended environments. Early work applied classical algorithms such as DQN to text-based games~\cite{dqn2015}. Recent policy optimization methods include GRPO~\citep{shao2024deepseekmath}, which uses group-based relative advantage estimation without a separate critic, and DAPO~\citep{yu2025dapo}, which improves stability through dynamic sampling and clip-higher strategies; both perform strongly on embodied tasks~\citep{ALFWorld2021}, information seeking~\citep{gaia2023}, and web navigation~\citep{webdancer2025}. All of them operate on single-episode trajectories, which limits their applicability where context resets occur. Our Multi-episode ARPO extends DAPO to trajectories of multiple episodes by propagating trajectory-level rewards across episode boundaries, so that checkpointing strategies themselves become learnable.

\noindent \textbf{Context Engineering in LLM Agents.}
Managing context is a central challenge for LLM agents, especially in long-horizon interactions where context windows are limited and irrelevant history accumulates. One line of work compresses context to stay within token budgets: Mem1~\citep{mem12025} applies compression after every function call, which controls length effectively but, on a rigid schedule, can discard useful information and disrupt high-level planning. CORAL instead lets the agent decide when compression is beneficial from the task state, so that compression happens when it is worth doing rather than on a fixed schedule. A second line processes fixed long inputs incrementally; MemAgent~\citep{memagent2025}, for example, targets a static long document. Agentic long-horizon settings differ in that context grows dynamically through interaction, and CORAL addresses that accumulation rather than a fixed input.


\section{Conclusion}

We address contextual bloat in LLM-driven agents, where accumulated environmental feedback and intermediate reasoning degrade long-horizon performance. CORAL equips agents with a callable toolset for managing their own working memory: a Memory Management tool that checkpoints task progress and verified facts, and a Context Optimization tool that resets context while preserving those checkpoints. Supervised Fine-Tuning instills these skills, and Multi-episode ARPO then propagates trajectory-level rewards across episode boundaries so the agent can discover effective policies. On GAIA, multi-hop Knowledge QA, and the harder BrowseComp, CORAL substantially outperforms existing methods while consuming roughly seven times fewer input tokens than the ReAct agent it replaces---the more consequential result, we think, since it shows that for long-horizon agents managing attention well is not a tax paid for accuracy but a source of it. Appendix~\ref{sec:limitations} discusses the framework's limitations and the research directions they open.

\section*{Acknowledgments}

This work was supported in part by the Research Grants Council under the Areas of Excellence scheme grant AoE/E-601/22-R.

\bibliography{colm2026_conference}

@book{kahneman1973attention,
  title={Attention and Effort},
  author={Kahneman, Daniel},
  year={1973},
  publisher={Prentice-Hall},
  address={Englewood Cliffs, NJ}
}

@incollection{baddeley1974working,
  title={Working memory},
  author={Baddeley, Alan D. and Hitch, Graham},
  booktitle={Psychology of Learning and Motivation},
  volume={8},
  pages={47--89},
  year={1974},
  publisher={Academic Press}
}

@article{sweller1988cognitive,
  title={Cognitive load during problem solving: Effects on learning},
  author={Sweller, John},
  journal={Cognitive Science},
  volume={12},
  number={2},
  pages={257--285},
  year={1988}
}

@article{risko2016cognitive,
  title={Cognitive offloading},
  author={Risko, Evan F. and Gilbert, Sam J.},
  journal={Trends in Cognitive Sciences},
  volume={20},
  number={9},
  pages={676--688},
  year={2016}
}

@article{gig2025,
  title={Group-in-group policy optimization for llm agent training},
  author={Feng, Lang and Xue, Zhenghai and Liu, Tingcong and An, Bo},
  journal={arXiv preprint arXiv:2505.10978},
  year={2025}
}

@article{timrun2025,
  title={Beyond Context Limits: Subconscious Threads for Long-Horizon Reasoning},
  author={Luo, Hongyin and Morgan, Nathaniel and Li, Tina and Zhao, Derek and Ngo, Ai Vy and Schroeder, Philip and Yang, Lijie and Ben-Kish, Assaf and O'Brien, Jack and Glass, James},
  journal={arXiv preprint arXiv:2507.16784},
  year={2025}
}

@article{memagent2025,
  title={MemAgent: Reshaping Long-Context LLM with Multi-Conv RL-based Memory Agent},
  author={Yu, Hongli and Chen, Tinghong and Feng, Jiangtao and Chen, Jiangjie and Dai, Weinan and Yu, Qiying and Zhang, Ya-Qin and Ma, Wei-Ying and Liu, Jingjing and Wang, Mingxuan and others},
  journal={arXiv preprint arXiv:2507.02259},
  year={2025}
}

@article{mem12025,
    author = {Zhou, Zijian and Qu, Ao and Wu, Zhaoxuan and Kim, Sunghwan and Prakash, Alok and Rus, Daniela and Zhao, Jinhua and Low, Bryan Kian Hsiang and Liang, Paul Pu},
    title = {MEM1: Learning to Synergize Memory and Reasoning for Efficient Long-Horizon Agents},
    journal = {arXiv preprint arXiv:2506.15841},
    year = {2025}
}

@inproceedings{gaia2023,
  title={Gaia: a benchmark for general ai assistants},
  author={Mialon, Gr{\'e}goire and Fourrier, Cl{\'e}mentine and Wolf, Thomas and LeCun, Yann and Scialom, Thomas},
  booktitle={The Twelfth International Conference on Learning Representations},
  year={2023}
}

@article{bc_en2025,
  title={Browsecomp: A simple yet challenging benchmark for browsing agents},
  author={Wei, Jason and Sun, Zhiqing and Papay, Spencer and McKinney, Scott and Han, Jeffrey and Fulford, Isa and Chung, Hyung Won and Passos, Alex Tachard and Fedus, William and Glaese, Amelia},
  journal={arXiv preprint arXiv:2504.12516},
  year={2025}
}

@inproceedings{dqn2015,
      title={Language Understanding for Text-based Games Using Deep Reinforcement Learning}, 
      author={Karthik Narasimhan and Tejas Kulkarni and Regina Barzilay},
      booktitle={Proceedings of the 2015 Conference on Empirical Methods in Natural Language Processing},
      year={2015}
}

@inproceedings{ALFWorld2021,
               title ={{ALFWorld: Aligning Text and Embodied
               Environments for Interactive Learning}},
               author={Mohit Shridhar and Xingdi Yuan and
               Marc-Alexandre C\^ot\'e and Yonatan Bisk and
               Adam Trischler and Matthew Hausknecht},
               booktitle = {Proceedings of the International
               Conference on Learning Representations (ICLR)},
               year = {2021}
}

@article{react2022,
  title={ReAct: Synergizing Reasoning and Acting in Language Models},
  author={Yao, Shunyu and Zhao, Jeffrey and Yu, Dian and Du, Nan and Shafran, Izhak and Narasimhan, Karthik and Cao, Yuan},
  journal={arXiv preprint arXiv:2210.03629},
  year={2022}
}

@misc{gpt4o,
  title = {Introducing OpenAI GPT-4o},
  author = {OpenAI},
  url = {https://openai.com/index/hello-gpt-4o/},
  year={2024}
}

@misc{claude4,
  title = {Introducing Claude 4},
  author = {Anthropic},
  url = {https://www.anthropic.com/news/claude-4},
  year={2025}
}

@article{r1,
  title={{DeepSeek-R1}: Incentivizing reasoning capability in {LLMs} via reinforcement learning},
  author={Guo, Daya and Yang, Dejian and Zhang, Haowei and Song, Junxiao and Zhang, Ruoyu and Xu, Runxin and Zhu, Qihao and Ma, Shirong and Wang, Peiyi and Bi, Xiao and others},
  journal={arXiv preprint arXiv:2501.12948},
  year={2025}
}

@misc{openaidr,
    title = {Deep Research System Card} ,
    author = {OpenAI},
    url = {https://cdn.openai.com/deep-research-system-card.pdf},
    year = {2025}
}

@article{r1searcher2025,
  title={R1-Searcher: Incentivizing the Search Capability in LLMs via Reinforcement Learning},
  author={Song, Huatong and Jiang, Jinhao and Min, Yingqian and Chen, Jie and Chen, Zhipeng and Zhao, Wayne Xin and Fang, Lei and Wen, Ji-Rong},
  journal={arXiv preprint arXiv:2503.05592},
  year={2025}
}

@article{webdancer2025,
  title={WebDancer: Towards Autonomous Information Seeking Agency},
  author={Wu, Jialong and Li, Baixuan and Fang, Runnan and Yin, Wenbiao and Zhang, Liwen and Tao, Zhengwei and Zhang, Dingchu and Xi, Zekun and Jiang, Yong and Xie, Pengjun and others},
  journal={arXiv preprint arXiv:2505.22648},
  year={2025}
}

@article{webthinker2025,
  author       = {Xiaoxi Li and
                  Jiajie Jin and
                  Guanting Dong and
                  Hongjin Qian and
                  Yutao Zhu and
                  Yongkang Wu and
                  Ji{-}Rong Wen and
                  Zhicheng Dou},
  title        = {WebThinker: Empowering Large Reasoning Models with Deep Research Capability},
  journal      = {arXiv preprint arXiv:2504.21776},
  year         = {2025}
}

@article{li2025torl,
  title={Torl: Scaling tool-integrated rl},
  author={Li, Xuefeng and Zou, Haoyang and Liu, Pengfei},
  journal={arXiv preprint arXiv:2503.23383},
  year={2025}
}

@article{yang2025llm,
  title={How Is LLM Reasoning Distracted by Irrelevant Context? An Analysis Using a Controlled Benchmark},
  author={Yang, Minglai and Huang, Ethan and Zhang, Liang and Surdeanu, Mihai and Wang, William and Pan, Liangming},
  journal={arXiv preprint arXiv:2505.18761},
  year={2025}
}

@article{qiao2024agent,
  title={Agent planning with world knowledge model},
  author={Qiao, Shuofei and Fang, Runnan and Zhang, Ningyu and Zhu, Yuqi and Chen, Xiang and Deng, Shumin and Jiang, Yong and Xie, Pengjun and Huang, Fei and Chen, Huajun},
  journal={Advances in Neural Information Processing Systems},
  volume={37},
  pages={114843--114871},
  year={2024}
}

@misc{gemini,
  title = {Gemini Deep Research},
  author={Gemini},
  howpublished = {\url{https://gemini.google/overview/deep-research}},
  year={2025}
}

@article{shinn2023reflexion,
  title={Reflexion: Language agents with verbal reinforcement learning},
  author={Shinn, Noah and Cassano, Federico and Gopinath, Ashwin and Narasimhan, Karthik and Yao, Shunyu},
  journal={Advances in Neural Information Processing Systems},
  volume={36},
  pages={8634--8652},
  year={2023}
}

@article{liu2025scalecua,
  title={ScaleCUA: Scaling Open-Source Computer Use Agents with Cross-Platform Data},
  author={Liu, Zhaoyang and Xie, JingJing and Ding, Zichen and Li, Zehao and Yang, Bowen and Wu, Zhenyu and Wang, Xuehui and Sun, Qiushi and Liu, Shi and Wang, Weiyun and others},
  journal={arXiv preprint arXiv:2509.15221},
  year={2025}
}

@article{erdogan2025plan,
  title={Plan-and-act: Improving planning of agents for long-horizon tasks},
  author={Erdogan, Lutfi Eren and Lee, Nicholas and Kim, Sehoon and Moon, Suhong and Furuta, Hiroki and Anumanchipalli, Gopala and Keutzer, Kurt and Gholami, Amir},
  journal={arXiv preprint arXiv:2503.09572},
  year={2025}
}

@article{huang2024understanding,
  title={Understanding the planning of LLM agents: A survey},
  author={Huang, Xu and Liu, Weiwen and Chen, Xiaolong and Wang, Xingmei and Wang, Hao and Lian, Defu and Wang, Yasheng and Tang, Ruiming and Chen, Enhong},
  journal={arXiv preprint arXiv:2402.02716},
  year={2024}
}

@article{wu2025gui,
  title={GUI-Reflection: Empowering Multimodal GUI Models with Self-Reflection Behavior},
  author={Wu, Penghao and Ma, Shengnan and Wang, Bo and Yu, Jiaheng and Lu, Lewei and Liu, Ziwei},
  journal={arXiv preprint arXiv:2506.08012},
  year={2025}
}

@article{wu2025resum,
  title={ReSum: Unlocking Long-Horizon Search Intelligence via Context Summarization},
  author={Wu, Xixi and Li, Kuan and Zhao, Yida and Zhang, Liwen and Ou, Litu and Yin, Huifeng and Zhang, Zhongwang and Jiang, Yong and Xie, Pengjun and Huang, Fei and others},
  journal={arXiv preprint arXiv:2509.13313},
  year={2025}
}

@misc{kimi_researcher,
  title = {Kimi-Researcher: End-to-End RL Training for Emerging Agentic Capabilities},
  author={Kimi},
  howpublished = {\url{https://moonshotai.github.io/Kimi-Researcher/}},
  year={2025}
}

@article{qian2025toolrl,
  title={Toolrl: Reward is all tool learning needs},
  author={Qian, Cheng and Acikgoz, Emre Can and He, Qi and Wang, Hongru and Chen, Xiusi and Hakkani-T{\"u}r, Dilek and Tur, Gokhan and Ji, Heng},
  journal={arXiv preprint arXiv:2504.13958},
  year={2025}
}

@article{wang2025otc,
  title={Otc: Optimal tool calls via reinforcement learning},
  author={Wang, Hongru and Qian, Cheng and Zhong, Wanjun and Chen, Xiusi and Qiu, Jiahao and Huang, Shijue and Jin, Bowen and Wang, Mengdi and Wong, Kam-Fai and Ji, Heng},
  journal={arXiv e-prints},
  pages={arXiv--2504},
  year={2025}
}

@article{shao2024deepseekmath,
  title={Deepseekmath: Pushing the limits of mathematical reasoning in open language models},
  author={Shao, Zhihong and Wang, Peiyi and Zhu, Qihao and Xu, Runxin and Song, Junxiao and Bi, Xiao and Zhang, Haowei and Zhang, Mingchuan and Li, YK and Wu, Yang and others},
  journal={arXiv preprint arXiv:2402.03300},
  year={2024}
}

@article{yu2025dapo,
  title={Dapo: An open-source llm reinforcement learning system at scale},
  author={Yu, Qiying and Zhang, Zheng and Zhu, Ruofei and Yuan, Yufeng and Zuo, Xiaochen and Yue, Yu and Dai, Weinan and Fan, Tiantian and Liu, Gaohong and Liu, Lingjun and others},
  journal={arXiv preprint arXiv:2503.14476},
  year={2025}
}

@article{zhang2025nemotron,
  title={Nemotron-research-tool-n1: Tool-using language models with reinforced reasoning},
  author={Zhang, Shaokun and Dong, Yi and Zhang, Jieyu and Kautz, Jan and Catanzaro, Bryan and Tao, Andrew and Wu, Qingyun and Yu, Zhiding and Liu, Guilin},
  journal={arXiv preprint arXiv:2505.00024},
  year={2025}
}

@article{xia2025agentrm,
  title={Agentrm: Enhancing agent generalization with reward modeling},
  author={Xia, Yu and Fan, Jingru and Chen, Weize and Yan, Siyu and Cong, Xin and Zhang, Zhong and Lu, Yaxi and Lin, Yankai and Liu, Zhiyuan and Sun, Maosong},
  journal={arXiv preprint arXiv:2502.18407},
  year={2025}
}

@article{li2025websailor,
  title={WebSailor: Navigating Super-human Reasoning for Web Agent},
  author={Li, Kuan and Zhang, Zhongwang and Yin, Huifeng and Zhang, Liwen and Ou, Litu and Wu, Jialong and Yin, Wenbiao and Li, Baixuan and Tao, Zhengwei and Wang, Xinyu and others},
  journal={arXiv preprint arXiv:2507.02592},
  year={2025}
}

@article{tao2025webshaper,
  title={Webshaper: Agentically data synthesizing via information-seeking formalization},
  author={Tao, Zhengwei and Wu, Jialong and Yin, Wenbiao and Zhang, Junkai and Li, Baixuan and Shen, Haiyang and Li, Kuan and Zhang, Liwen and Wang, Xinyu and Jiang, Yong and others},
  journal={arXiv preprint arXiv:2507.15061},
  year={2025}
}

@article{jin2025search,
  title={Search-r1: Training llms to reason and leverage search engines with reinforcement learning},
  author={Jin, Bowen and Zeng, Hansi and Yue, Zhenrui and Yoon, Jinsung and Arik, Sercan and Wang, Dong and Zamani, Hamed and Han, Jiawei},
  journal={arXiv preprint arXiv:2503.09516},
  year={2025}
}

@article{fang2025cognitive,
  title={Cognitive kernel-pro: A framework for deep research agents and agent foundation models training},
  author={Fang, Tianqing and Zhang, Zhisong and Wang, Xiaoyang and Wang, Rui and Qin, Can and Wan, Yuxuan and Ma, Jun-Yu and Zhang, Ce and Chen, Jiaqi and Li, Xiyun and others},
  journal={arXiv preprint arXiv:2508.00414},
  year={2025}
}

@article{kwiatkowski2019natural,
  title={Natural questions: a benchmark for question answering research},
  author={Kwiatkowski, Tom and Palomaki, Jennimaria and Redfield, Olivia and Collins, Michael and Parikh, Ankur and Alberti, Chris and Epstein, Danielle and Polosukhin, Illia and Devlin, Jacob and Lee, Kenton and others},
  journal={Transactions of the Association for Computational Linguistics},
  volume={7},
  pages={453--466},
  year={2019},
  publisher={MIT Press One Rogers Street, Cambridge, MA 02142-1209, USA journals-info~…}
}

@inproceedings{yang2018hotpotqa,
  title={HotpotQA: A dataset for diverse, explainable multi-hop question answering},
  author={Yang, Zhilin and Qi, Peng and Zhang, Saizheng and Bengio, Yoshua and Cohen, William and Salakhutdinov, Ruslan and Manning, Christopher D},
  booktitle={Proceedings of the 2018 conference on empirical methods in natural language processing},
  pages={2369--2380},
  year={2018}
}

@article{ho2020constructing,
  title={Constructing a multi-hop qa dataset for comprehensive evaluation of reasoning steps},
  author={Ho, Xanh and Nguyen, Anh-Khoa Duong and Sugawara, Saku and Aizawa, Akiko},
  journal={arXiv preprint arXiv:2011.01060},
  year={2020}
}

@article{trivedi2022musique,
  title={MuSiQue: Multihop Questions via Single-hop Question Composition},
  author={Trivedi, Harsh and Balasubramanian, Niranjan and Khot, Tushar and Sabharwal, Ashish},
  journal={Transactions of the Association for Computational Linguistics},
  year={2022}
}

@inproceedings{press2023measuring,
  title={Measuring and narrowing the compositionality gap in language models},
  author={Press, Ofir and Zhang, Muru and Min, Sewon and Schmidt, Ludwig and Smith, Noah A and Lewis, Mike},
  booktitle={Findings of the Association for Computational Linguistics: EMNLP 2023},
  pages={5687--5711},
  year={2023}
}

@article{li2024numinamath,
  title={Numinamath: The largest public dataset in ai4maths with 860k pairs of competition math problems and solutions},
  author={Li, Jia and Beeching, Edward and Tunstall, Lewis and Lipkin, Ben and Soletskyi, Roman and Huang, Shengyi and Rasul, Kashif and Yu, Longhui and Jiang, Albert Q and Shen, Ziju and others},
  journal={Hugging Face repository},
  volume={13},
  number={9},
  pages={9},
  year={2024}
}

@article{moshkov2025aimo2,
  title   = {AIMO-2 Winning Solution: Building State-of-the-Art Mathematical Reasoning Models with OpenMathReasoning dataset},
  author  = {Ivan Moshkov and Darragh Hanley and Ivan Sorokin and Shubham Toshniwal and Christof Henkel and Benedikt Schifferer and Wei Du and Igor Gitman},
  year    = {2025},
  journal = {arXiv preprint arXiv:2504.16891}
}

@article{li2025search,
  title={Search-o1: Agentic search-enhanced large reasoning models},
  author={Li, Xiaoxi and Dong, Guanting and Jin, Jiajie and Zhang, Yuyao and Zhou, Yujia and Zhu, Yutao and Zhang, Peitian and Dou, Zhicheng},
  journal={arXiv preprint arXiv:2501.05366},
  year={2025}
}

@article{zheng2025deepresearcher,
  title={Deepresearcher: Scaling deep research via reinforcement learning in real-world environments},
  author={Zheng, Yuxiang and Fu, Dayuan and Hu, Xiangkun and Cai, Xiaojie and Ye, Lyumanshan and Lu, Pengrui and Liu, Pengfei},
  journal={arXiv preprint arXiv:2504.03160},
  year={2025}
}

@article{sun2025simpledeepsearcher,
  title={SimpleDeepSearcher: Deep information seeking via web-powered reasoning trajectory synthesis},
  author={Sun, Shuang and Song, Huatong and Wang, Yuhao and Ren, Ruiyang and Jiang, Jinhao and Zhang, Junjie and Bai, Fei and Deng, Jia and Zhao, Wayne Xin and Liu, Zheng and others},
  journal={arXiv preprint arXiv:2505.16834},
  year={2025}
}
\bibliographystyle{colm2026_conference}

\appendix
\appendix

\section{Training Details}
\label{sec:training-details}

In this section, we provide detailed hyperparameters for both the Supervised Fine-Tuning (SFT) and Reinforcement Learning (RL) stages.

\subsection{Supervised Fine-Tuning}

We perform full-parameter fine-tuning on Qwen3-8B using the following hyperparameters:

\begin{table}[ht]
    \centering
    \caption{Hyperparameters for Supervised Fine-Tuning.}
    \label{tab:sft-hyperparams}
    \small
    \begin{tabular}{l c}
    \toprule
    \textbf{Hyperparameter} & \textbf{Value} \\
    \midrule
    Number of epochs & 3 \\
    Learning rate & 1e-5 \\
    Per-device batch size & 4 \\
    Gradient accumulation steps & 2 \\
    Effective batch size & 64 \\
    Warmup ratio & 0.05 \\
    Max sequence length & 65,536 \\
    Precision & bfloat16 \\
    Optimizer & AdamW (DeepSpeed ZeRO-2) \\
    RoPE scaling & YaRN (factor=2.0) \\
    \bottomrule
    \end{tabular}
\end{table}

\subsection{Reinforcement Learning}

We apply Multi-episode ARPO (ME-ARPO) on the SFT checkpoint with the following hyperparameters:

\begin{table}[ht]
    \centering
    \caption{Hyperparameters for Reinforcement Learning.}
    \label{tab:rl-hyperparams}
    \small
    \begin{tabular}{l c}
    \toprule
    \textbf{Hyperparameter} & \textbf{Value} \\
    \midrule
    Actor learning rate & 1e-6 \\
    Train batch size & 64 \\
    Mini-batch size & 16 \\
    Samples per prompt ($n$) & 8 \\
    Clip ratio (low) & 0.2 \\
    Clip ratio (high) & 0.28 \\
    Temperature (rollout) & 1.0 \\
    Top-p (rollout) & 1.0 \\
    Max turns per trajectory & 50 \\
    KL regularization & None \\
    \bottomrule
    \end{tabular}
\end{table}

\section{Evaluation Metrics and Statistical Analysis}
\label{sec:eval-metrics}

In this section, we provide a detailed description of the evaluation metrics used in our experiments and report additional statistical analysis including standard deviations for the GAIA benchmark.

\subsection{Evaluation Metrics}

Due to the inherent stochasticity in language model generation, we follow common practice in evaluating language agents by sampling multiple trajectories per question and reporting aggregated metrics. Given $k$ independent runs for each question, we define the following metrics:

\noindent \textbf{Pass@1 (Avg@$k$)} computes the average accuracy across $k$ independent runs:
\begin{equation}
\text{Pass@1 (Avg@}k\text{)} = \frac{1}{n} \sum_{i=1}^{n} \left( \frac{1}{k} \sum_{j=1}^{k} \mathbb{I}[\text{correct}_{i,j}] \right)
\end{equation}
where $n$ is the number of questions and $\mathbb{I}[\text{correct}_{i,j}]$ indicates whether the $j$-th run for question $i$ is correct. This metric represents the expected accuracy when deploying the model in production with a single inference.

\noindent \textbf{Pass@$k$} measures whether at least one of the $k$ runs produces a correct answer:
\begin{equation}
\text{Pass@}k = \frac{1}{n} \sum_{i=1}^{n} \mathbb{I}\left[\exists j \in [1,k]: \text{correct}_{i,j}\right]
\end{equation}
This metric reflects the upper bound of model capability given multiple attempts, useful for assessing the model's potential when combined with verification mechanisms.

\noindent \textbf{Maj@$k$} (Majority Voting) selects the most frequent answer among $k$ runs as the final prediction:
\begin{equation}
\text{Maj@}k = \frac{1}{n} \sum_{i=1}^{n} \mathbb{I}\left[\text{majority}(\{a_{i,j}\}_{j=1}^{k}) = a_i^*\right]
\end{equation}
where $a_{i,j}$ is the answer from the $j$-th run for question $i$, and $a_i^*$ is the ground truth. This metric leverages the wisdom of crowds to improve robustness.

\noindent \textbf{Best Pass@1 and Worst Pass@1.} When conducting multiple independent evaluation sessions (each with a single run per question), we also report the best and worst pass@1 across sessions:
\begin{align}
\text{Best Pass@1} &= \max_{s \in \mathcal{S}} \left( \frac{1}{n} \sum_{i=1}^{n} \mathbb{I}[\text{correct}_{i}^{(s)}] \right) \\
\text{Worst Pass@1} &= \min_{s \in \mathcal{S}} \left( \frac{1}{n} \sum_{i=1}^{n} \mathbb{I}[\text{correct}_{i}^{(s)}] \right)
\end{align}
where $\mathcal{S}$ denotes the set of evaluation sessions. These metrics characterize the variance in model performance across different runs, which is particularly important for understanding deployment reliability.

\subsection{Statistical Analysis on GAIA}

To provide a comprehensive view of model performance, we report the mean and standard deviation of Pass@1 across 8 independent runs on the GAIA benchmark in~\autoref{tab:gaia-std}. The standard deviation captures the run-to-run variance, which is crucial for understanding the reliability of the model in practical deployments.

\begin{table}[ht]
    \centering
    \caption{Detailed performance statistics on GAIA (8 runs). We report Mean $\pm$ Std for Pass@1, along with Best Pass@1, Worst Pass@1, Maj@8, and Pass@8.}
    \label{tab:gaia-std}
    \small
    \renewcommand{\arraystretch}{1.1}
    \begin{tabular}{l c c c c}
    \toprule
    \textbf{Model} & \textbf{L1} & \textbf{L2} & \textbf{L3} & \textbf{Avg} \\
    \midrule
    \multicolumn{5}{c}{\cellcolor{gray!20}\textsc{Qwen3-8B + SFT}} \\
    \midrule
    Pass@1 (Avg@8) & 47.1 $\pm$ 4.1 & 43.5 $\pm$ 4.5 & 15.6 $\pm$ 9.4 & 41.6 $\pm$ 3.5 \\
    Best Pass@1 & 53.8 & 51.9 & 33.3 & 47.6 \\
    Worst Pass@1 & 41.0 & 38.5 & 8.3 & 36.9 \\
    Maj@8 & 51.3 & 50.0 & 16.7 & 46.6 \\
    Pass@8 & 74.4 & 78.9 & 33.3 & 71.8 \\
    \midrule
    \multicolumn{5}{c}{\cellcolor{gray!20}\textsc{Qwen3-8B + RL}} \\
    \midrule
    Pass@1 (Avg@8) & 46.2 $\pm$ 3.9 & 45.7 $\pm$ 3.7 & 19.8 $\pm$ 9.9 & 42.8 $\pm$ 1.9 \\
    Best Pass@1 & 51.3 & 51.9 & 33.3 & 45.6 \\
    Worst Pass@1 & 41.0 & 42.3 & 0.0 & 39.8 \\
    Maj@8 & 48.7 & 46.2 & 25.0 & 44.7 \\
    Pass@8 & 74.4 & 73.1 & 33.3 & 68.9 \\
    \bottomrule
    \end{tabular}
\end{table}

Several observations emerge from this analysis:
\begin{itemize}[leftmargin=1em]
    \item \textbf{Lower variance with RL}: The RL model exhibits notably lower standard deviation on the overall accuracy (1.9\%) compared to the SFT model (3.5\%), suggesting that RL training leads to more consistent performance.
    \item \textbf{High variance on Level 3}: Both models show substantial variance on Level 3 tasks ($\sim$9-10\%), reflecting the challenging nature of long-horizon reasoning tasks where small differences in early decisions can lead to divergent outcomes.
    \item \textbf{Gap between Pass@8 and Avg@8}: The significant gap between Pass@8 and Avg@8 (e.g., 71.8\% vs. 41.6\% for SFT) indicates room for improvement through better answer selection or verification mechanisms.
\end{itemize}

\section{Data Preparation Walkthrough}
\label{sec:data-walkthrough}

This appendix follows the data pipeline end to end: how questions are synthesized, how a trajectory is generated, and how it becomes training instances. The point we most want to establish is negative: at no stage do we \textit{stitch} together separate reasoning processes. Every multi-episode trajectory is one model solving one question in one uninterrupted run, and the multi-episode structure is discovered in that run rather than imposed on it.

\definecolor{tracebg}{RGB}{248,248,248}
\definecolor{traceframe}{RGB}{180,180,180}
\definecolor{tracetitlebg}{RGB}{85,85,85}

\newtcblisting{traceenv}[1]{
    enhanced, breakable,
    colback=tracebg,
    colframe=traceframe,
    coltitle=white,
    colbacktitle=tracetitlebg,
    fonttitle=\bfseries\small,
    title={#1},
    left=5pt, right=5pt, top=4pt, bottom=4pt,
    boxrule=0.4pt, arc=1.5pt,
    listing only,
    listing options={
        basicstyle=\ttfamily\fontsize{7.5}{9.5}\selectfont,
        breaklines=true,
        breakautoindent=false,
        columns=fullflexible,
        keepspaces=true,
        showstringspaces=false,
        aboveskip=0pt,
        belowskip=0pt,
    }
}

\subsection{Question synthesis with WebShaper}
\label{sec:webshaper}

Because our training questions are synthesized rather than collected, we summarize the WebShaper pipeline~\citep{tao2025webshaper} for readers unfamiliar with it. The distinguishing idea is to fix the \textit{formal structure} of a question before any web content is retrieved, which inverts the usual order and makes difficulty a controllable parameter instead of an emergent property of whatever pages happened to be crawled.

Each information-seeking task is represented as a set of \textit{Knowledge Projections}: set-theoretic constraints over entities and relations, so that ``the company founded in the same year as event $X$'' is a constraint rather than a string. The pipeline then runs in two stages. First, seed questions are constructed from random walks over Wikipedia and filtered by executing agent rollouts, retaining only questions that at least one rollout answers correctly---this discards questions that are unanswerable or ill-posed rather than merely hard. Second, an Expander agent performs \textit{agentic expansion}: it takes a seed question's formal representation, retrieves and validates new web information, and layer-wise replaces leaf constants with sub-questions that identify them indirectly. Expansion increases the number of hops while explicitly checking two failure modes that plague naively generated multi-hop data---redundant constraints that add length without adding reasoning, and shortcuts that let the answer be inferred from a single shallow clue.

The result is a question distribution whose reasoning depth we can dial, which matters for our purposes because CORAL's benefits only appear once the horizon is long enough for context clutter to bind. Questions are generated over public encyclopedic entities and are independent of every benchmark we evaluate on.

\subsection{Stage 1: Trajectory generation}

A commercial model (Claude Sonnet 4) receives a synthesized question together with the CORAL system prompt (Appendix~\ref{sec:system-prompt}) and all four tools, and runs to completion. It decides on its own when to call MM and when to call CO; we do not intervene. A representative trajectory, abridged, looks as follows.

\begin{traceenv}{A generated two-episode trajectory}
[Episode 1]
  System prompt + Question
  -> Think:       Plan the research...
  -> Action:      MM(ADD task_progress: "Step 1: search for X; Step 2: find Y")
  -> Observation: Memory updated
  -> Think:       Let me search...
  -> Action:      WebSearch("query about X")
  -> Observation: Search results...
  -> Think:       Found X = 42. Let me save this.
  -> Action:      MM(ADD verified_fact: "X = 42")
  -> Observation: Memory updated
  -> Think:       Context is getting long, and step 1 is done. Let me optimize.
  -> Action:      CO(reason: "completed step 1, ready for step 2")
  -> Observation: [Context reset -- memory state returned]

[Episode 2]                       <- begins automatically at the CO boundary
  System prompt + Question + Memory state written by Episode 1
  -> Think:       I know X = 42 from my memory. Now I need to find Y...
  -> Action:      WebSearch("query about Y")
  -> ...
  -> Think:       Final answer is X + Y = 42 + 58 = 100.
  -> Answer:      100
\end{traceenv}

Coherence across the boundary is structural rather than something we have to enforce: Episode~2's opening context is built from the memory state that Episode~1 wrote immediately before calling CO, so the agent resumes knowing exactly which sub-tasks are done and which facts are established. This is also why the system prompt makes synchronizing \texttt{Task Progress} a mandatory precondition for invoking CO (Appendix~\ref{sec:system-prompt}); a CO call issued with a stale memory would indeed produce an incoherent successor episode.

\subsection{Stage 2: Episode splitting}

For SFT we cut each trajectory at its CO boundaries, and each resulting episode becomes an independent training conversation:

\begin{itemize}[leftmargin=1em]
    \item \textbf{Episode 1} $\;\rightarrow\;$ input \texttt{[System + Question]}, target \texttt{[Think, MM, Search, MM, Think, CO]}
    \item \textbf{Episode 2} $\;\rightarrow\;$ input \texttt{[System + Question + Memory state]}, target \texttt{[Think, Search, ..., Answer]}
\end{itemize}

The essential detail is that Episode~2's input is \textit{not} a truncation of the raw history. It is the system prompt, the original question, and the memory state returned by the CO tool, assembled with the context reset template of Appendix~\ref{sec:context-reset-template}---byte for byte what the agent will be handed at inference time when it invokes CO. Splitting is therefore not a convenience but a correctness requirement: presenting both episodes as one flat conversation would train the model on a context it will never see.

This operation is also the sole reason the conversation count exceeds the trajectory count. Our 3,173 raw trajectories contain 509 CO invocations in total, each of which contributes one additional conversation, giving the 3,682 SFT conversations reported in Section~\ref{sec:exp-settings} and \autoref{tab:ablation}.

At inference time the same boundary logic runs in reverse. When the agent emits a CO call, the system clears the conversation, rebuilds the context from the system prompt, the original question, and the current memory state, and continues generation. Training-time and inference-time episode boundaries are thus produced by one mechanism, which is what allows a model trained on split episodes to behave correctly on unsplit tasks.

\section{Working Memory Behavior and Robustness}
\label{sec:memory-robustness}

\subsection{Tool usage by outcome}

\autoref{tab:tool-usage} breaks the aggregate tool-usage statistics of \autoref{tab:efficiency} down by whether the trajectory ended up correct.

\begin{table}[ht]
    \centering
    \caption{MM and CO invocations per trajectory on GAIA, split by outcome. Across all three configurations, trajectories that fail invoke CO roughly twice as often as trajectories that succeed.}
    \label{tab:tool-usage}
    \small
    \renewcommand{\arraystretch}{1.1}
    \begin{tabular}{l l c c}
    \toprule
    \textbf{Configuration} & \textbf{Outcome} & \textbf{MM / traj.} & \textbf{CO / traj.} \\
    \midrule
    \multirow{3}{*}{Prompting-only} & Correct   & 20.11 & 2.05 \\
                                    & Incorrect & 20.69 & 4.80 \\
                                    & Overall   & 20.58 & 4.29 \\
    \midrule
    \multirow{3}{*}{SFT}            & Correct   & 3.37  & 0.12 \\
                                    & Incorrect & 4.44  & 0.29 \\
                                    & Overall   & 3.93  & 0.21 \\
    \midrule
    \multirow{3}{*}{RL}             & Correct   & 3.30  & 0.12 \\
                                    & Incorrect & 4.70  & 0.28 \\
                                    & Overall   & 4.06  & 0.21 \\
    \bottomrule
    \end{tabular}
\end{table}

The by-outcome split sharpens the reading we gave in Section~\ref{sec:efficiency}. In every configuration, failing trajectories reset about twice as often as succeeding ones---0.12 versus 0.28--0.29 after training, 2.05 versus 4.80 before it. The natural causal story runs from difficulty to resets rather than the reverse: a trajectory that is going badly wanders, accumulates clutter, and triggers the conditions the system prompt lists as reset-worthy. Reset frequency is therefore better read as a symptom of struggle than as a lever to be maximized, which is consistent with training reducing CO usage by a factor of twenty while raising accuracy by more than twenty points. MM usage falls by a similar factor and for the same reason: the untrained model re-writes essentially the same memory entry after nearly every step, whereas the trained model writes a smaller number of well-chosen checkpoints.

\subsection{Sensitivity to memory corruption}
\label{sec:memory-corruption}

Because the agent treats its working memory as authoritative, a natural question is what happens when that memory is wrong. We probed this by manually corrupting entries in the memory store at different points in otherwise-successful GAIA trajectories and observing whether the agent noticed. The study is qualitative---we report the failure modes we observed rather than a corruption-rate curve---but the boundary it reveals is sharp and follows directly from what information remains available to the agent.

\begin{itemize}[leftmargin=1em]
    \item \textbf{Corruption immediately after planning is absorbed silently.} When we altered the \texttt{task\_progress} entry right after the agent recorded its initial plan, the agent adopted the corrupted plan without objection and pursued it to completion. At that point in the trajectory there is nothing to check the plan against: no evidence has been gathered yet, so a plausible-looking plan is indistinguishable from the one the agent actually wrote.
    \item \textbf{Mid-trajectory corruption is frequently caught.} When we altered a \texttt{verified\_fact} several reasoning steps after it had been established, the agent often detected the inconsistency, because the search results and page contents that originally produced the fact were still present in the conversation. Detection was not reliable, but the agent had the means to detect it.
    \item \textbf{Post-reset corruption is undetectable in principle.} When we altered a fact that had been written before a CO invocation, the agent never recovered. The reset had already discarded the supporting evidence, so no amount of subsequent reasoning could surface the contradiction.
\end{itemize}

We draw a design conclusion from the middle and last cases. The conversation history is not merely redundant with the memory store; it functions as a cross-validation channel, and CO closes that channel permanently for everything the memory currently holds. This is the concrete reason CORAL is asymmetric between its two tools. MM sharpens the agent's focus on key information while leaving the history intact and is therefore cheap to use liberally, whereas CO trades away the agent's ability to audit itself and should be reserved for contexts that have genuinely become unmanageable. The trained models' behavior---3.9 MM calls against 0.21 CO calls per trajectory---matches this prescription, which we find reassuring, though we note it was learned from demonstrations rather than explicitly optimized for. Appendix~\ref{sec:limitations} discusses provenance-carrying memory as a way to close the gap.

\section{Limitations and Future Work}
\label{sec:limitations}

We discuss three limitations of CORAL and the research directions each one opens.

\noindent \textbf{Credit assignment across episodes.} Multi-episode ARPO broadcasts the trajectory reward equally to every episode (\autoref{eq:reward_propagation}). This is deliberately the simplest viable scheme, and it sits in tension with our own motivation: if a trajectory mixes productive groundwork with failed detours, rewarding both identically discards exactly the distinction the framework is built around. We adopted it because our aim in adding the SFT and RL stages was to establish whether dedicated training could improve CORAL on open-weight models at all---which it does, from 20.4\% under prompting to 41.6\% after SFT and 42.8\% after RL---and given that focus we did not explore more sophisticated credit assignment designs across episode boundaries.

The obvious refinement may nonetheless be the wrong one. Discounting earlier episodes is standard practice in long-horizon RL, where early actions often have diminishing causal influence on the final outcome. That reasoning does not transfer cleanly to CORAL, because an episode boundary here is not an arbitrary point in time: it is created by a CO call, a deliberate and learned decision by the agent to consolidate progress and start fresh. Earlier episodes are therefore not earlier attempts that may have gone wrong; they are structured preparatory phases that perform task decomposition, fact discovery, and verification through MM, and that directly enable the later episodes to succeed. Penalizing them would discourage precisely the behavior CORAL is trying to install.

The deeper obstacle is that moving beyond equal broadcasting requires a principled way to assess the quality of an individual episode---whether it contributed verified facts that later episodes actually used, or whether it explored unproductive directions. Designing such episode-level quality metrics is a non-trivial research problem in its own right, and we regard it as an independent question deserving dedicated study rather than a knob to be swept. Two directions seem most concrete: episode-level reward models trained on trajectory outcomes, and contribution-weighted schemes based on how heavily an episode's checkpointed facts are used downstream. We leave a systematic comparison to future work.

\noindent \textbf{Trust in the memory store.} CORAL's working memory is authoritative by construction: nothing in the framework lets the agent doubt a checkpointed fact. Appendix~\ref{sec:memory-corruption} shows this is a real failure mode with a sharp boundary determined by what evidence remains available. A fact corrupted while the conversation still contains the search results that produced it is frequently caught; a fact corrupted after a CO reset never is. The conversation history therefore functions as a cross-validation channel, and CO closes that channel permanently for everything the memory currently holds.

This is the strongest argument we know for CORAL's asymmetry between its two tools---MM preserves history and is cheap to use liberally, CO trades away the agent's ability to audit itself---and the trained models' behavior matches the prescription. But it is a mitigation, not a solution. The natural next step is provenance-aware memory: each verified fact carries a pointer to the evidence that established it, so that after a reset the agent can re-fetch and re-verify rather than trusting a bare assertion. This also matters in deployment, since a fact checkpointed from a low-quality source currently propagates through the remainder of a trajectory with no record of where it came from.

\noindent \textbf{Evaluation scale.} GAIA's text-only validation split contains 12 Level~3 questions. This is the standard split used throughout this literature, and we use it for comparability, but it is small enough that a single question shifts the Level~3 score by 8.3 points; the corresponding standard deviations we measure are 9--10\% (\autoref{tab:gaia-std}). Our Level~3 numbers should therefore be read as directional evidence for a trend, not as precise estimates. We mitigate this in three ways: averaging 8 independent samples per question rather than reporting a single run, reporting the full distribution including the worst run, and corroborating on five Knowledge QA benchmarks and on BrowseComp, all with larger test sets. The pattern that CORAL's advantage widens with task difficulty is consistent across all of them, which is the claim we actually rely on.

Two further scope limits are worth stating plainly. Our BrowseComp evaluation covers the first 150 questions with a single proprietary backbone, chosen for cost; a full-benchmark evaluation across backbones would be a stronger test. And while the prompting-only framework is validated on five backbones (Appendix~\ref{sec:backbone}), the SFT and RL stages are validated on Qwen3-8B, with one supporting SFT run on Qwen2.5-7B. Extending the training pipeline to more model families is the most direct way to strengthen the paper's central claim.

\section{CORAL Across Backbones}
\label{sec:backbone}

\autoref{tab:backbone} reports the full cross-backbone study summarized in Section~\ref{sec:additional-eval}. Every row uses CORAL as a prompting-time wrapper only, with no SFT and no RL, so the comparison isolates what the framework contributes to a model that has never been trained to use it.


\begin{table}[ht]
    \centering
    \small
    \renewcommand{\arraystretch}{1.1}
    \begin{tabularx}{\linewidth}{
    >{\arraybackslash}p{3.5cm}
    >{\arraybackslash}p{2.5cm}
    >{\centering\arraybackslash}X
    >{\centering\arraybackslash}X
    >{\centering\arraybackslash}X
    }
    \toprule
    \textbf{Backbone} & \textbf{Benchmark} & ReAct & \cellcolor{blue!15}+\,CORAL & $\Delta$ \\
    \midrule
    Qwen2.5-7B-Instruct & GAIA & 17.5 & \cellcolor{blue!15}13.6 & $-$3.9 \\
    Qwen3-8B            & GAIA & 19.4 & \cellcolor{blue!15}20.4 & $+$1.0 \\
    QwQ-32B             & GAIA & 26.2 & \cellcolor{blue!15}32.0 & $+$5.8 \\
    Qwen3-32B           & GAIA & 34.0 & \cellcolor{blue!15}36.3 & $+$2.3 \\
    Claude Sonnet 4     & GAIA & 60.2 & \cellcolor{blue!15}65.0 & $+$4.8 \\
    \midrule
    Claude Sonnet 4     & BrowseComp$^{\dagger}$ & 22.7 & \cellcolor{blue!15}30.7 & $+$8.0 \\
    \bottomrule
    \end{tabularx}
    \caption{CORAL used purely as a prompting-time wrapper, with no training of any kind, across five backbones. The benefit grows with the capability of the backbone and is largest on the benchmark that stresses long-horizon browsing hardest. $^{\dagger}$The first 150 BrowseComp questions; both systems see exactly the same questions and the same web search and browse tools, and differ only in whether the MM and CO tools are available.}
    \label{tab:backbone}
\end{table}

Two caveats on reading this table. First, the GAIA rows for Qwen3-8B and Claude Sonnet 4 are the same numbers as the corresponding rows of \autoref{tab:gaia-results}, reproduced here so that the trend across backbones can be read in one place. Second, the $\Delta$ column understates what training recovers at the low end: Qwen2.5-7B-Instruct loses 3.9 points under prompting-only CORAL, but the same backbone reaches 34.4\% after SFT on our data, against a 17.5\% ReAct baseline. The prompting-only deficit measures a tool-calling deficiency, not a ceiling on the framework.

\section{BrowseComp Case Study}
\label{sec:browsecomp-case}

To make concrete what the 8.0-point BrowseComp gap in \autoref{tab:backbone} consists of, we trace a single question on which CORAL succeeds and the ReAct baseline fails, and on which the difference is directly attributable to Memory Management. Both runs use Claude Sonnet 4 with identical search and browse tools; the only difference is whether MM and CO are available.

\begin{tcolorbox}[enhanced, breakable, colback=gray!4, colframe=gray!55,
                  coltitle=white, colbacktitle=gray!55, fonttitle=\bfseries\small,
                  title={BrowseComp Question \#6}, boxrule=0.4pt, arc=1.5pt,
                  left=6pt, right=6pt, top=5pt, bottom=5pt]
\small
\textit{The headquarters for a company widely reported to be founded in the same year as a prominent worker's strike related to print media, though it is sometimes reported to have been founded two years prior, moved back to the downtown area of a city in 2014. A nu metal band, where the band name is an acronym, was formed in this city. Two years after the formation of this band, a different band with a name inspired by a mythological character released a song about loving food made with insects which references a product produced by the aforementioned company. An artist posted a sketch on their blog, 20 years after the formation of the nu metal band, where they stated they may have been listening to the song about loving food made with insects when they made the sketch. The artist graduated from school 7 years prior to posting the sketch. What is the full name of the school from which they graduated for this degree?}

\vspace{3pt}
\noindent\textbf{Ground truth:} Savannah College of Art and Design.
\end{tcolorbox}

This is a six-hop chain in which every hop consumes the output of the previous one, so a single misidentification invalidates everything downstream---the structure that makes context management decisive.

\subsection{CORAL: 36 turns, 7 MM checkpoints, correct}

CORAL decomposed the chain and committed each link to working memory as soon as it was verified. \autoref{tab:browsecomp-mm} lists the seven MM calls in the order they were made; each row corresponds to one hop of the chain.

\begin{table}[ht]
    \centering
    \small
    \renewcommand{\arraystretch}{1.15}
    \begin{tabularx}{\linewidth}{
    >{\centering\arraybackslash}p{0.4cm}
    >{\arraybackslash}p{2.9cm}
    >{\arraybackslash}X
    }
    \toprule
    \textbf{\#} & \textbf{MM checkpoint} & \textbf{Verified fact recorded} \\
    \midrule
    1 & Task decomposition & Objective and sub-task plan \\
    2 & Strike identified  & Newsboys' strike of 1899, New York \\
    3 & Company identified & Bumble Bee Seafoods (founded 1897); moved back to downtown San Diego in 2014 \\
    4 & Nu metal band      & P.O.D.\ (``Payable on Death''), formed in San Diego, 1992 \\
    5 & Band and song      & Mephiskapheles; ``Bumble Bee Tuna Song'' (1994), which references Bumble Bee products \\
    6 & Artist identified  & Joanna Davidovich; blog post of 10 June 2012 ($=1992+20$) mentioning the song \\
    7 & School confirmed   & BFA in Animation from SCAD, 2005 ($=2012-7$) \\
    \bottomrule
    \end{tabularx}
    \caption{The seven Memory Management calls CORAL issued while solving BrowseComp Question~\#6, in order. Each fact was checkpointed at the moment it was confirmed, which let the agent advance to the next hop without carrying forward the search results that had produced the previous one.}
    \label{tab:browsecomp-mm}
\end{table}

\subsection{ReAct: 47 turns, 42 searches, gave up}

The baseline committed to a wrong hypothesis on its fifth turn and never recovered:

\begin{itemize}[leftmargin=1.6em]
    \item \textbf{Turn 5.} Identified the company as Coca-Cola (founded 1892, Atlanta) rather than Bumble Bee Seafoods, and therefore fixed the city as Atlanta.
    \item \textbf{Turns 7--9.} Following from Atlanta, identified Sevendust as the nu metal band.
    \item \textbf{Turns 10--46.} Searched for a mythologically-named band that had released a song about insect food referencing Coca-Cola---a combination that does not exist---for more than thirty consecutive turns.
    \item \textbf{Turn 47.} Gave up, reporting that it had not ``been able to definitively identify the specific band''.
\end{itemize}

\subsection{Why the memory made the difference}

It is not that ReAct lacked the information: the correct company was retrievable by exactly the same searches. What ReAct lacked was any representation of \textit{which} of its beliefs the following thirty turns depended on. The Coca-Cola identification was made once, on turn 5, and thereafter existed only as one paragraph among forty-two search results; when the third hop began failing, nothing directed the agent's attention back to the premise rather than to the query. CORAL's MM calls make that dependency structure explicit, so a downstream dead end prompts re-examination of a short list of checkpointed entities rather than of an undifferentiated transcript.

\autoref{tab:browsecomp-tools} shows how differently the two agents then behave. CORAL issues fewer than half as many searches but nearly three times as many page browses: checkpointed intermediate results let it target specific pages instead of probing with queries. ReAct substitutes search volume for structure, and most of that volume explores consequences of a premise it never revisited.

\begin{table}[ht]
    \centering
    \small
    \renewcommand{\arraystretch}{1.15}
    \begin{tabular}{l c c}
    \toprule
    \textbf{Metric} & \textbf{CORAL} & \textbf{ReAct} \\
    \midrule
    Total turns    & 36 & 47 \\
    Web searches   & 17 & 42 \\
    Page browses   & 11 & 4 \\
    MM checkpoints & 7  & -- \\
    \midrule
    Outcome        & \textbf{correct} & gave up \\
    \bottomrule
    \end{tabular}
    \caption{Tool usage on BrowseComp Question~\#6.}
    \label{tab:browsecomp-tools}
\end{table}

\section{Comparison of Policy Optimization Objectives}
\label{sec:comparison-objectives}

In this section, we provide a detailed comparison of the policy optimization objectives used in GRPO, DAPO, and our proposed Multi-episode ARPO. The key differences are highlighted in \textcolor{red}{red}.

\noindent \textbf{GRPO (Group Relative Policy Optimization):}
\begin{equation}
\begin{aligned}
\mathcal{J}_{\text{GRPO}}(\theta) &= \mathbb{E}_{q\sim \mathcal{D}, \textcolor{red}{\{o_i\}_{i=1}^G}\sim \pi_{\theta_\text{old}}} \\
&\Bigg[ \textcolor{red}{\frac{1}{G}\sum_{i=1}^{G} \frac{1}{|o_i|}\sum_{t=1}^{|o_i|}}
\min \Big( r_{i,t}(\theta) \textcolor{red}{\hat{A}_{i,t}},  
\text{clip} \Big( r_{i,t}(\theta), 1 - \textcolor{red}{\varepsilon}, 1 + \textcolor{red}{\varepsilon} \Big) \textcolor{red}{\hat{A}_{i,t}} \Big) \quad - \textcolor{red}{\beta D_{\text{KL}}(\pi_{\theta} \| \pi_{\text{ref}})} \Bigg]
\end{aligned}
\end{equation}
where
\begin{equation}
r_{i,t}(\theta)=\frac{\pi_{\theta}(\textcolor{red}{o_{i,t}} \mid \textcolor{red}{q, o_{i,<t}})}{\pi_{\theta_{\text{old}}}(\textcolor{red}{o_{i,t}} \mid \textcolor{red}{q, o_{i,<t}})},\quad \textcolor{red}{\hat{A}_{i,t}} = \frac{R_i - \text{mean}(\{R_k\}_{k=1}^G)}{\text{std}(\{R_k\}_{k=1}^G)}
\end{equation}

\noindent \textbf{DAPO (Decouple Clip and Dynamic Sampling Policy Optimization):}
\begin{equation}
\begin{aligned}
\mathcal{J}_{\text{DAPO}}(\theta) &= \mathbb{E}_{q\sim \mathcal{D}, \textcolor{red}{\{o_i\}_{i=1}^G}\sim \pi_{\theta_\text{old}}} \\
&\Bigg[ \textcolor{red}{\frac{1}{\sum_{i=1}^G |o_i|}\sum_{i=1}^{G}\sum_{t=1}^{|o_i|}}
\min \Big( r_{i,t}(\theta) \textcolor{red}{\hat{A}_{i,t}},  
\text{clip} \Big( r_{i,t}(\theta), 1 - \textcolor{red}{\varepsilon_{\text{low}}}, 1 + \textcolor{red}{\varepsilon_{\text{high}}} \Big) \textcolor{red}{\hat{A}_{i,t}} \Big) \Bigg] \\
\text{s.t.} \quad &0 < \left|\{o_i \mid \texttt{is\_equivalent}(a, o_i)\}\right| < G
\end{aligned}
\end{equation}
where
\begin{equation}
r_{i,t}(\theta)=\frac{\pi_{\theta}(\textcolor{red}{o_{i,t}} \mid \textcolor{red}{q, o_{i,<t}})}{\pi_{\theta_{\text{old}}}(\textcolor{red}{o_{i,t}} \mid \textcolor{red}{q, o_{i,<t}})},\quad \textcolor{red}{\hat{A}_{i,t}} = \frac{R_i - \text{mean}(\{R_k\}_{k=1}^G)}{\text{std}(\{R_k\}_{k=1}^G)}
\end{equation}

\noindent \textbf{Multi-episode ARPO (Ours):}
\begin{equation}
\begin{aligned}
\mathcal{J}_{\text{ME-ARPO}}(\theta) &= \mathbb{E}_{q\sim \mathcal{D}, \textcolor{red}{\{\mathcal{T}_i\}_{i=1}^G}\sim \pi_{\theta_\text{old}}} \\
&\Bigg[\textcolor{red}{\frac{1}{\sum_{e \in \mathcal{E}}|e|}\sum_{e_i^j \in \mathcal{E}}\sum_{t=1}^{|e_i^j|}}
\min \Big( r_{i,j,t}(\theta) \textcolor{red}{\hat{A}_{i}^{j}},  
\text{clip} \Big( r_{i,j,t}(\theta), 1 - \varepsilon_{\text{low}}, 1 + \varepsilon_{\text{high}} \Big) \textcolor{red}{\hat{A}_{i}^{j}} \Big) \Bigg]
\end{aligned}
\end{equation}
where $\mathcal{E} = \{e_i^j \mid i \in [1,G], j \in [1, N_i]\}$ denotes the set of all episodes from all trajectories, and
\begin{equation}
r_{i,j,t}(\theta)=\frac{\pi_{\theta}(\textcolor{red}{e_{i,t}^j} \mid \textcolor{red}{c_i^j, e_{i,<t}^j})}{\pi_{\theta_{\text{old}}}(\textcolor{red}{e_{i,t}^j} \mid \textcolor{red}{c_i^j, e_{i,<t}^j})},\quad \textcolor{red}{\hat{A}_{i}^{j}} = \frac{R_i - \text{mean}(\mathcal{R})}{\text{std}(\mathcal{R})}
\end{equation}
where $\mathcal{R} = \{R_{i,j} \mid i \in [1,G], j \in [1, N_i]\}$ is the multiset of episode rewards for question $q$.

\begin{algorithm}[t]
\caption{Multi-Episode Agentic Reinforcement Policy Optimization (ME-ARPO)}
\label{alg:me-arpo}
\begin{algorithmic}[1]
\REQUIRE Initial policy model $\pi_{\theta}$; reward function $R(\cdot)$; task prompts with tool environment $\mathcal{D}$; hyperparameters $\varepsilon_{\text{low}}, \varepsilon_{\text{high}}$; group size $G$; number of training steps $M$; mini-batch iterations $\mu$
\FOR{step $= 1, \ldots, M$}
    \STATE Sample a batch $\mathcal{D}_b$ from $\mathcal{D}$
    \STATE Update the old policy model $\pi_{\theta_{\text{old}}} \gets \pi_{\theta}$
    \FOR{each question $q \in \mathcal{D}_b$}
        \STATE Sample $G$ multi-episode trajectories $\{\mathcal{T}_i^{(q)}\}_{i=1}^G \sim \pi_{\theta_{\text{old}}}(\cdot | q)$
        \STATE Compute trajectory-level rewards $\{R_i^{(q)}\}_{i=1}^G$ via $R(\cdot)$
        \STATE Propagate rewards to episodes: assign $R_i^{(q)}$ to each episode $e_i^j \in \mathcal{T}_i^{(q)}$
        \STATE Collect all episode rewards for question $q$: $\mathcal{R}^{(q)} = \{R_{i,j}^{(q)} \mid i \in [1,G], j \in [1, N_i]\}$
        \STATE Compute per-question statistics over episodes: $\bar{R}^{(q)} = \text{mean}(\mathcal{R}^{(q)})$, $\sigma^{(q)} = \text{std}(\mathcal{R}^{(q)})$
        \FOR{each episode $e_i^j$ in question $q$}
            \STATE Compute advantage: $\hat{A}_{i}^{j} = (R_{i,j}^{(q)} - \bar{R}^{(q)}) / \sigma^{(q)}$
        \ENDFOR
    \ENDFOR
    \STATE Collect all episodes $\mathcal{E} = \{(e_i^j, \hat{A}_{i}^{j}) \mid q \in \mathcal{D}_b, i \in [1, G], j \in [1, N_i]\}$
    \FOR{iteration $= 1, \ldots, \mu$}
        \STATE Update policy $\pi_{\theta}$ by maximizing the ME-ARPO objective $\mathcal{J}_{\text{ME-ARPO}}(\theta)$ using episodes in $\mathcal{E}$
    \ENDFOR
\ENDFOR
\ENSURE Optimized policy $\pi_{\theta}$
\end{algorithmic}
\end{algorithm}

\noindent \textbf{Key Differences:}
\begin{itemize}[leftmargin=1em]
    \item \textbf{GRPO $\rightarrow$ DAPO}: (1) Removes the KL divergence regularization term; (2) Adopts asymmetric clipping bounds ($\varepsilon_{\text{low}}, \varepsilon_{\text{high}}$) instead of symmetric clipping; (3) Changes normalization from per-output averaging to global token averaging; (4) Use Dynamic Sampling.
    \item \textbf{DAPO $\rightarrow$ ME-ARPO}: (1) Samples multi-episode trajectories $\{\mathcal{T}_i^{(q)}\}$ instead of single-episode outputs $\{o_i\}$; (2) Pools all episodes $e_i^j \in \mathcal{E}$ from all trajectories for batch optimization; (3) Propagates trajectory-level rewards to all episodes within each trajectory, then computes advantage $\hat{A}_{i}^{j}$ using per-question normalization statistics over all episodes, ensuring all episodes within a trajectory share the same advantage signal; (4) Conditions the probability ratio on episode-specific context $c_i^j$ rather than the original question $q$.
\end{itemize}

\section{Prompt Templates}
\label{sec:prompt-templates}

In this section, we provide the complete prompt templates used in our system, including the system prompt, tool definitions, the context reset template, and the LLM-as-a-Judge prompt.

\definecolor{promptbg}{RGB}{248,248,248}
\definecolor{promptframe}{RGB}{180,180,180}
\definecolor{titlebg}{RGB}{85,85,85}

\newtcblisting{promptenv}[1]{
    enhanced, breakable,
    colback=promptbg,
    colframe=promptframe,
    coltitle=white,
    colbacktitle=titlebg,
    fonttitle=\bfseries\small,
    title={#1},
    left=5pt, right=5pt, top=4pt, bottom=4pt,
    boxrule=0.4pt, arc=1.5pt,
    listing only,
    listing options={
        basicstyle=\ttfamily\fontsize{7.5}{9.5}\selectfont,
        breaklines=true,
        breakautoindent=false,
        columns=fullflexible,
        keepspaces=true,
        showstringspaces=false,
        aboveskip=0pt,
        belowskip=0pt,
    }
}

\subsection{System Prompt}
\label{sec:system-prompt}

The system prompt instructs the agent to operate in an adaptive research loop with structured working memory management.

\begin{promptenv}{System Prompt}
You are an advanced AI research assistant. Your mission is to deliver comprehensive, accurate, and well-structured answers by systematically executing research tasks.

# GUIDING PRINCIPLES
- **Rigor:** Verify information by consulting multiple primary sources.
- **Synthesis:** Intelligently connect new findings with established knowledge to form a coherent understanding.
- **Adaptability:** Recognize that the initial plan is a hypothesis. Continuously refine it based on new information. Do not adhere to a flawed plan.
- **Focus:** Prioritize actions and data that directly advance the current task. Discard irrelevant information.

# OPERATIONAL PROTOCOL: THE ADAPTIVE RESEARCH LOOP
You will operate in a continuous cycle of execution and reflection until the research goal is met.

## 1. Deconstruct & Plan
Briefly analyze the user's request and formulate an initial, high-level plan as a sequence of logical sub-tasks. Store this plan in your working memory. This initial plan is considered a draft and is expected to evolve.

## 2. Execute One Sub-Task
Execute the *single, highest-priority* sub-task from your current plan using the available tools. Focus on completing this one step thoroughly.

### Tool Usage Rules
- Conduct searches step by step: search multiple attributes of single entity separately, process multiple entities one by one.
- Snippets in search results contain only partial information; browse the full page to ensure you don't miss relevant content.
- You **MUST** use the browser tool to access all URLs, whether from user input or search results.
- Cross-validate information by accessing multiple URLs from search results.
- Actively explore valuable links to deepen your research.

## 3. Reflect & Adapt
After every tool use or execution step, you **MUST** pause to perform this reflection sequence:

1.  **Analyze Outcome:** Review the results of your last action. What new information was gathered? Was the action successful? Did it yield unexpected results?
2.  **Update Progress:** Immediately update your working memory.
    - **MANDATORY.** Update your internal task list to reflect what was just completed and what the immediate next step is. This ensures your progress is never lost.
3.  **Evaluate & Adjust Plan:** Compare the new information against your overall plan. Ask yourself:
    - Is the current plan still the most effective path?
    - Did the recent findings open up a better approach or reveal a flaw in my strategy?
    - Should the priority of the remaining sub-tasks be changed?
4.  **Adapt:** Based on your evaluation, **modify the plan in your working memory if necessary.** This may involve re-ordering, adding, or removing sub-tasks.

## 4. Synthesize & Answer
Once you determine that all refined sub-tasks are complete, perform a final review of all `Verified Facts` units in your working memory. Synthesize this information into a single, comprehensive, and well-structured final answer.

# WORKING MEMORY PROTOCOL
You will maintain a structured working memory to track the plan, store key findings, and ensure focus.

## Memory Structure
Your working memory consists of two mandatory components:

1.  **Task Progress**: A dynamic, structured overview of the research plan and the current state of the research.
2.  **Verified Facts**: A collection of atomic, confirmed pieces of information. Each fact should be a concise statement, directly relevant to the research goal, and ideally supported by a source URL.

## Memory Operations & Rules
You can use the `memory-management-tool` tool to manage your working memory.

* **Always Maintain the Plan:** You **MUST** proactively update the task progress.
* **Store Critical Findings:** When you discover new, crucial information, immediately add it as a new verified fact. Do not wait.

# CONTEXT MANAGEMENT PROTOCOL
To maintain focus and operational efficiency, you MUST proactively manage the conversational context. A cluttered context degrades performance. The `context_optimization` tool is your mechanism for pruning irrelevant history and preserving only essential information.

## Core Principle: State Preservation
Your working memory, specifically the `Task Progress` unit, is your state. The primary goal of context management is to condense the conversational history *without losing your state*.

## Mandatory Pre-Optimization Action
Before **EVERY** use of the `context-optimization-tool` tool, you **MUST** perform the following state-saving action:

1.  **Synchronize Working Memory:** Ensure your `Task Progress` unit is perfectly up-to-date. It MUST reflect:
    - All sub-tasks that have just been completed.
    - The **precise and immediate** next sub-task you intend to execute.

*Rationale: This action guarantees that your critical plan and progress are preserved through the optimization, allowing for a seamless continuation of your task.*

## Strategic Triggers for Optimization
Invoke the `context-optimization-tool` tool in the following situations:

-   **Routine Pruning:** After a batch of 3-5 tool invocations (especially information-gathering tools like `search` or `browse`). This prevents gradual context bloat.
-   **Post-Reflection Milestone:** After completing a major sub-task and performing the "Reflect & Adapt" step. This is a natural point to clean the slate before starting the next major phase.
-   **Error Recovery / Dead End:** If you detect circular reasoning, repeated failed attempts, or have pursued a dead end, use this sequence:
    1.  **Reflect** on the failure.
    2.  **Adapt** the plan to correct the strategy.
    3.  **Optimize** the context to discard the failed path's history and start fresh on the new, corrected path.
\end{promptenv}

\subsection{Tool Definitions}
\label{sec:tool-definitions}

We equip the agent with four tools. Below we provide the name, description, and parameter schema for each tool.

\begin{promptenv}{Tool 1: web-search-tool}
Description: Search web pages using search engine. Use for obtaining latest information or finding references.

Parameters:
{
  "query": {
    "type": "array",
    "items": {"type": "string"},
    "description": "Array of search query strings to execute simultaneously. Each search query should be in Google search style, using 2-5 keywords."
  }
}
\end{promptenv}

\begin{promptenv}{Tool 2: browse-webpage-tool}
Description: Intelligently browse and analyze web pages to extract goal-specific information.

This tool:
- Retrieves webpage content from any accessible URLs
- Analyzes content based on your specified goal/purpose
- Extracts relevant text, summaries, and useful links
- Handles multiple URLs simultaneously (up to 5)

Use this tool when you need to:
- Research information from specific websites
- Extract data from online articles or documentation
- Gather insights from multiple web sources
- Find relevant links for further exploration

The tool will return structured information including key findings, summaries, and relevant external links.

Parameters:
{
  "browse_items": {
    "type": "array",
    "items": {
      "url":  {"type": "string", "description": "The webpage URL to analyze."},
      "goal": {"type": "string", "description": "The specific information or insights you want to extract from the webpage."}
    }
  }
}
\end{promptenv}

\begin{promptenv}{Tool 3: memory-management-tool}
Description: Manage your working memory to maintain task continuity.

This tool allows you to:
- ADD: Store critical knowledge units (task progress, verified facts)
- UPDATE: Specify the existing knowledge unit to update by corresponding `unit_id`
- DELETE: Specify the existing knowledge unit to delete by corresponding `unit_id`

MANDATORY:
The content of `Task Progress` must be a single YAML object with the following keys:

```yaml
objective: "A concise statement of the user's ultimate goal."
tasks:
  - description: "A clear, actionable description of the sub-task."
    status: "PENDING | COMPLETED | FAILED"
```

Parameters:
{
  "operations": [{
    "operation": {"type": "string", "enum": ["ADD", "UPDATE", "DELETE"]},
    "knowledge_unit": {
      "type":      {"type": "string", "description": "task_progress | verified_fact"},
      "content":   {"type": "string", "description": "Concise, actionable description"},
      "relevance": {"type": "string", "description": "Why this is important"},
      "source":    {"type": "array",  "description": "Source references"}
    },
    "unit_id":   {"type": "string", "description": "ID for UPDATE/DELETE"},
    "rationale": {"type": "string", "description": "Why this operation is necessary"}
  }]
}
\end{promptenv}

\begin{promptenv}{Tool 4: context-optimization-tool}
Description: Perform the optimization of the conversational context to maintain focus and operational efficiency.

Parameters:
{
  "optimization_reason": {
    "type": "string",
    "description": "Detailed explanation of why context optimization is needed now"
  },
  "task_checkpoint": {
    "type": "string",
    "description": "Detailed description of current task state and what needs to be done next"
  }
}
\end{promptenv}

\subsection{Context Reset Template}
\label{sec:context-reset-template}

When the agent invokes the \texttt{context-optimization-tool}, the system triggers a context reset. The agent's conversation history is cleared and replaced with the following template, which preserves the agent's working memory and task state.

\begin{promptenv}{Context Reset Template (New Episode Prompt)}
Your conversation history has been cleared, but all essential information is preserved in your working memory below.

<user_task>
{user_query}
</user_task>

<current_working_memory>
{memory_summary}
</current_working_memory>

<task_checkpoint>
{task_checkpoint}
</task_checkpoint>

<next_steps>
1. Review your working memory and task checkpoint to understand current task state
2. Continue execution from where you left off
</next_steps>

Remember: Your working memory contains all the critical information you saved. Trust it and continue forward without repeating completed work.
\end{promptenv}

\subsection{LLM-as-a-Judge Prompt}
\label{sec:llm-judge-prompt}

We use an LLM-as-a-Judge approach to evaluate answer correctness when exact string matching is insufficient (\eg open-ended QA tasks). The judge model receives the following system prompt.

\begin{promptenv}{LLM-as-a-Judge System Prompt}
You are a highly specialized AI assistant acting as an expert evaluator. Your sole purpose is to determine if the **final conclusion** of a "Submitted Answer" is correct by comparing it to the "Official Correct Answer".

**Your Task:**

You will be given three pieces of information:
1.  `Question`: The original problem or query.
2.  `Official Correct Answer`: The definitive, correct answer to the question.
3.  `Submitted Answer`: The answer provided by a user or another system, which may include intermediate reasoning and a final conclusion.

Your task is to evaluate the Submitted Answer based on one single criterion:

**Does the final conclusion of the Submitted Answer exactly match the Official Correct Answer?**

**Critical Instructions:**

1.  **Focus Exclusively on the Final Conclusion:** You must **ignore** the intermediate steps, reasoning, methodology, or any explanations provided in the Submitted Answer. The correctness, elegance, or flaws in the reasoning process are completely irrelevant to your evaluation. Your judgment is based only on the final result.

2.  **Definition of "Exact Match":**
    *   **Numerical Answers:** The numbers must be mathematically equivalent. For example, if the official answer is `0.5`, a submitted answer of `1/2` or `50%` is considered a match. An answer of `0.501` is not.
    *   **Textual Answers:** The text must be semantically identical. Minor differences in wording are acceptable only if the meaning is precisely the same.
    *   **Mathematical Formulas:** The formulas must be mathematically equivalent.
    *   **Sets or Lists:** The elements must be the same, though the order does not matter unless specified by the question.

3.  **Output Format:** Your response must be a JSON object containing two keys: `judgment` and `reason`.
    *   `judgment`: (String) Must be either `"Correct"` or `"Incorrect"`.
    *   `reason`: (String) A brief, one-sentence explanation for your judgment.
        *   If correct, state that the final conclusion matches.
        *   If incorrect, state what the submitted conclusion was and how it differs from the official answer.

**Example:**
*   **Question:** "What is the capital of Japan?"
*   **Official Correct Answer:** "Tokyo"
*   **Submitted Answer:** "Based on my research, Japan has many famous cities like Osaka and Kyoto, but the capital is Tokyo."
*   **Output:**
    ```json
    {
      "judgment": "Correct",
      "reason": "The final conclusion 'Tokyo' in the Submitted Answer exactly matches the Official Correct Answer."
    }
    ```
\end{promptenv}

\end{document}